\documentclass{thesis.cs.pub.ro}
\usepackage{blindtext}
\usepackage{tabulary}
\makeindex
\begin{document}

\title{A Local - Global Approach to
Semantic Segmentation in Aerial Images}
\author{Alina - Elena Marcu}
\date{2016}

\titleen{A Local - Global Approach to
Semantic Segmentation in Aerial Images}
\titlero{O abordare duală a segmentării semantice în imagini aeriene}
\adviser{Assoc.Prof. Marius Leordeanu}
\titlefooteren{Bucharest, 2016}
\titlefooterro{București, 2016}


\newcommand{\project}{MySuperProject}

\begin{frontmatter} 

\begin{titlepage}
	\begin{center}
		{\Large University ``Politehnica'' of Bucharest}
		\par\vspace*{2mm}
		{\Large Automatic Control and Computers Faculty,
		
		Computer Science and Engineering Department}
		\par\vspace*{3mm}
		\begin{table*}[h]
        	\begin{center}
				\begin{tabular}{cccc}
                    \includegraphics[width=0.13\textwidth]{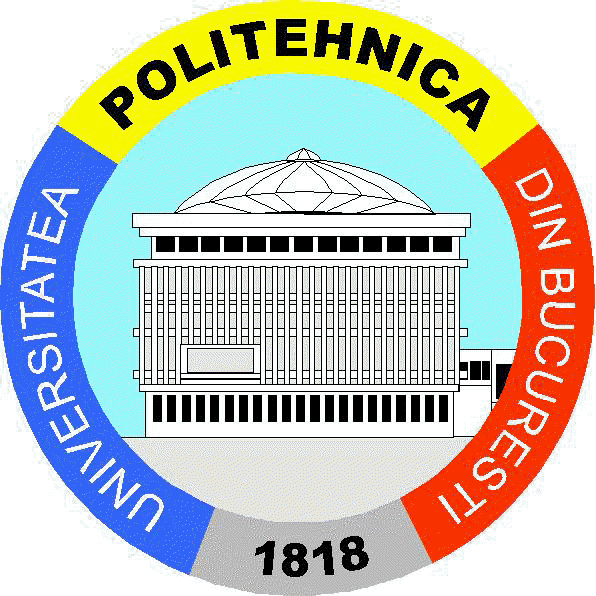}
					& & &
					\includegraphics[width=0.30\textwidth]{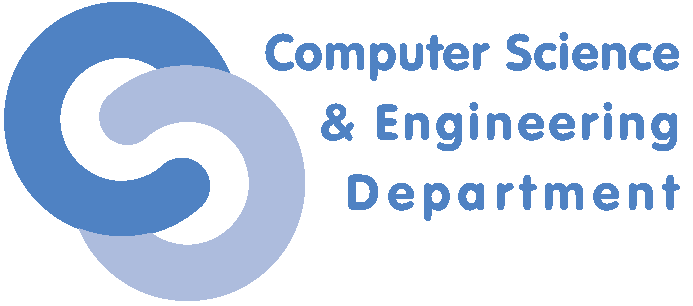}
            	\end{tabular}
			\end{center}
		\end{table*}
		
		\par\vspace*{20mm}
		{\Huge MASTER THESIS}
		\par\vspace*{15mm}
		{\Huge \VARtitleen }
		\par\vspace*{20mm}
		\begin{table*}[h]
        	\begin{center}
				\begin{tabular}{lcccccl}
					\Large \textbf{\Large Scientific Advisor:}
					\vspace*{1mm} &&&&&& \Large \textbf{\Large Author:}\vspace*{1mm} \\
					\Large \VARadviser &&&&&& \Large \VARauthor
				\end{tabular}
			\end{center}
		\end{table*}

		\par\vspace*{40mm}
		\Large \VARtitlefooteren
	\end{center}
\end{titlepage}

\begin{acknowledgements}
\vspace*{7cm}
\begin{center}
I want to thank my supervisor, Associate Prof. Marius Leordeanu, for his deep insights and guidance throughout these years. His patience and encouragement inspired my passion for research.
\end{center}

\end{acknowledgements}

\begin{abstract}

Visual context is important in object recognition and it is still an open problem in computer vision. Along with the advent of deep convolutional neural networks, using contextual information with such systems starts to receive attention in the literature. At the same time, aerial imagery is gaining momentum. While advances in deep learning make good progress in aerial image analysis, this problem still poses many great challenges. Aerial images are often taken under poor lighting conditions and contain low resolution objects, many times occluded by trees or taller buildings. In this domain, in particular, visual context could be of great help, but there are still very few papers that consider context in aerial image understanding. In this thesis we introduce context as a complementary way of recognizing objects. We propose a dual - stream deep neural network model that processes information along two independent pathways, one for local and another for global visual reasoning. The two are later combined in the final layers of processing. Our model learns to combine local object appearance as well as information from the larger scene in a complementary way, such that together they form a powerful classifier. We test our dual - stream network on the task of segmentation of buildings and roads in aerial images and obtain state-of-the-art results on the Massachusetts Buildings Dataset. We also introduce two new datasets, for buildings and roads segmentation, respectively, and study the relative importance of local appearance versus the larger scene, as well as their performance
in combination. We also extend the segmentation task to other classes that we find in aerial imagery, namely meadows, forest and water. While our local - global model could also be useful in general recognition tasks, we clearly demonstrate the effectiveness of visual context in conjunction with deep nets for aerial image understanding.

\end{abstract}


\tableofcontents
\newpage

\listoffigures
\newpage

\listoftables
\newpage

\printabbrev

\end{frontmatter} 


\chapter{Introduction}
\label{chapter:introduction}

Humans possess the ability of detecting and locating regular objects in the scene as well as recognizing novel objects. Despite the difficulty of the scene and the variation in color, texture, form, scale, different viewing points of these objects, some of them partially obstructed from view and so on, humans are capable of recognizing objects with little effort on a daily basis. Therefore the human visual system is an inspiration for building an object recognition system to emulate it. Based only on their appearance, objects have certain patterns that make us differentiate them from one another. These patterns represent an arrangement of features or descriptors. In some cases, local information may seem insufficient for real - world detection problems. Aerial imagery is one such case and the goal of this thesis is to develop machine learning techniques that tackle this problem.

The  field  of  UAVs  is  enjoying  a  great  increase  in  interest  nowadays  in  research,  in  industry, basically in all aspects of technology, especially in the last decade. Aerial images offer a new and exciting research  direction  as  unmanned  aerial  vehicles  are  beginning  to  have  increasingly  more  commercial success. A  novel  and  challenging  idea  is  the  combination  of  drones  and  computer  vision,  enabling  unmanned aerial vehicles to have a certain degree of understanding of the overflown area.  Some application examples in which smart vision will improve the way drones fly are Automatic Feature  Matching  Recognition  and  Image-based  control  that  will  enable  localization,  mapping  and navigation in real-time, in the absence of GPS. Efficient semantic interpretation of the scene from above will  provide  a  higher  level  of  understanding  in  order  to  perform  tasks  such  as  monitoring  the environment  or  finding  a  safe  place  to  land, object  recognition  and  3D  inference  methods  using machine learning techniques that will significantly improve obstacle avoidance, planning and safe flying.

The process of aerial image interpretation implies aerial image examination with the sole purpose of identifying various discriminative characteristics of the objects of interest. In order to obtain total scene understanding from an aerial image several steps are needed. Given an image,  a  segmentation  step  is  applied  in  order  to  divide  the  scene  into  regions  of  certain  categories (such as residential areas, flood, forest, roads etc.), basically see the whole environment as a fully interconnected place of all categories, interacting with each other. The  regions  of  interest  can  be  further  used  in  the  process  of  individual  object  detection.

The thesis addresses the problem of aerial image interpretation as a pixel labeling task. Given an RGB image as the one presented in \labelindexref{Figure}{img:fig_introduction} in the first column, the main goal is to assign each pixel in the image a label with the class it belongs to. In the second column, the image shows the result of the semantic segmentation, where three types of labels are assigned (red – roads, blue – others, green – houses).

\fig[scale=0.95]{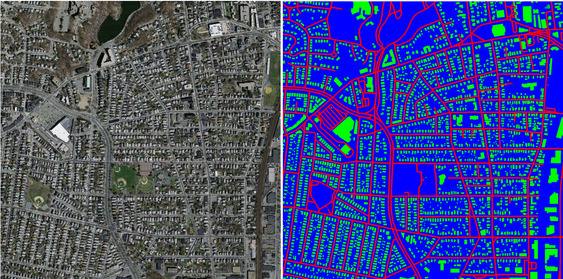}{img:fig_introduction}{RGB aerial image example (left) and corresponding ground truth labels (right) for roads (red), others (blue) and buildings (green).}

Object recognition in aerial imagery is enjoying a growing interest today, due to the recent advancements in computer vision and deep learning, along with important improvements in low - cost high performance GPUs. The possibility of accurately recognizing different types of objects in aerial images, such as buildings, roads, vegetation and other categories, could greatly help in many applications, such as creating and keeping up-to-date maps, improving urban planning, environment monitoring and disaster relief. Besides the practical need for accurate aerial image interpretation systems, this domain also offers specific scientific challenges to the computer vision domain. Aerial images require the recognition of very small objects, seen from above under difficult lighting conditions, which are sometimes occluded or only partially seen. One point we make in this thesis is that visual context is vital for accurate recognition in such cases.

Building extraction systems from aerial images have applications in a wide range of areas and offer a different perspective over the object recognition  problem.  Disaster relief is just one such example of real life application. Such a demanding problem requires real - time response of which humans are incapable of doing manually. The research that we have conducted focuses on identifying efficient methods for buildings recognition  from  top - down  view  aerial  images and  developing  an  efficient  automatic  system  capable of identifying individual buildings. We start with the task of building segmentation from aerial images and then expand our application on other classes found in such imagery.

We study the importance of visual spatial context and propose a dual - stream deep convolutional neural network (DCNN) that combines local appearance with global scene information in a complementary way. Thus, the object is seen both as a separate entity from the perspective of its own appearance, but also as a part of a larger scene, which acts as its complement and implicitly contains information about it. Our local - global model offers a dual view of the object, with one processing pathway, based on a state-of-the-art deep CNN that focuses on local, object level information, and a second one, which considers information from the larger area around the object of interest. The two pathways are then joined into a final subnet composed of three fully - connected layers, where the intermediate results are combined and potential disagreements solved for a final output. We formulate the problem as one of segmentation in the sense of finding an accurate shape for the object of interest. The task of aerial image labeling is solved using our patch - based combined network trained jointly, end-to-end.

We bring the following contributions: 

\begin{enumerate}
\item We studied the relative importance of local appearance versus the larger scene, as well as their performance in combination.
\item We demonstrate experimentally that larger visual context is important for semantic segmentation in aerial images and show superior performance to current
state-of-the-art methods on the Massachusetts Buildings Dataset.
\item We propose a novel
dual - stream deep CNN architecture, with two independent processing pathways,
one for local and the other for global image interpretation. Also, demonstrating the importance of the larger visual scene context in aerial imagery is relevant, since current techniques in aerial imagery
focus only on local object appearance. 
\item We show in our experiments that the two pathways learn to process information complementarily in order to obtain an improved output.
\item Highlight the difficulty of semantic segmentation in aerial imagery. We trained detectors on several classes (buildings, roads, meadows, water and forest) and underline the challenges of detecting each object individually.
\end{enumerate}

The outline of our work presented in this master thesis is further detailed. In \labelindexref{Chapter}{chapter:related-work} we present the most relevant work for our problem starting from basic hand - crafted models up to the current state-of-the-art deep CNN for visual recognition. In \labelindexref{Chapter}{chapter:problem-description} we better formulate the problem of semantic segmentation in aerial images. We also provide the intuition behind our proposed method and the steps that motivate our design decisions.  In \labelindexref{Chapter}{chapter:experiments} we detail our experimental scenarios and test our theory on different segmentation tasks. We perform experiments on buildings detection on two large - scale datasets and report state-of-the-art results on one publicly available dataset. We extend our application to other classes and train different networks for segmentation of roads, meadows, forest and water concluding our work in \labelindexref{Chapter}{chapter:conclusions} with a summary of our main contribution to the field of semantic segmentation in aerial images and also presenting some future research directions in this matter.

Part of this work was integrated in an ArXiv paper and published at \cite{marcu2016dual}.

\chapter{Visual context and aerial imagery}
\label{chapter:related-work}

There are very few attempts to create an automated object detector for aerial images and none of  them  offer  satisfactory  results.  Therefore,  this  branch  of  research  is  relatively  new  and  not  yet enough  developed.  Given  the  complexity  imposed  by  the  object  recognition  problem,  several approaches  for  solving  this  particular  problem  were  proposed  over  the  time,  making  it  an  intensive research topic in computer vision. 

Different aspects of object representation, learning and recognition require  dividing  the  main  problem  into  smaller ones  and  for  each  and  every  one  of  them  a  diversity  of  methods  with  promising  results  was implemented. The diversity of the existing approaches is given by the different data sources, as well as the  learning  models  and  data  representation  schemes  used.  To  get  a  better  understanding  of  the process of image interpretation, a basic - level definition is mandatory. 

Most of the existing methods of object detection use intrinsic object features, capable of handling object transformations such as displacement, rotation or scaling, treating the object independently from the environment where it resides. Object detection in outdoor environments is a challenging problem given the various properties that  a  certain  type  of  object  holds,  such  as  resolution,  ambient  lighting,  size  and  shape,  orientation, color etc.  For example, in the case of building detection, depending on the area type from  which the object of interest  was provided, urban or rural areas, there is a high probability that the object of interest is labeled as a “building” for urban locations and  as  a  “house”  for  rural  location, with very few resemblance with each other.  The  context  surrounding  the  object  can  provide  even  more information and limit the ambiguity in recognizing an object. Context could play a fundamental role in aerial image understanding, especially in cases of poor resolution, poor lighting or occlusion. For example, a square in the middle of a residential area could be more confidently labeled as a building than in the middle of a road or a large body of water. Thus, the same square, with exactly the same appearance, could be seen differently. There  is  a  lot  of  relevant  work  for  various  computer  vision  problems  that  study  and  use contextual information.  Earlier  approaches used  global  scene  features  for  object  recognition  ~\cite{torralba2003contextual, oliva2007role, torralba2006contextual}. Other works used only the immediate neighbourhood of an object, which often provides strong cues for  image  recognition  or  tracking  in  video ~\cite{zhu2015segdeepm, collins2005online, leordeanu2016labeling}. 

Due to the spatial coherence of images, the labels of nearby image pixels tend to be highly correlated, dependent of one another in the form of contextual  relationships between objects, such as co-occurrences between different categories ~\cite{rabinovich2007objects}. The presence of different object detectors in the vicinity of the box of interest is also known to increase recognition performance ~\cite{felzenszwalb2010object}.  Other methods  based  on  relationships  between  objects  include  modeling spatial  relations  as  a structured  prediction  problem  ~\cite{desai2011discriminative}. Improving accuracy detection is obtained by exploiting structure in the output labels and incorporating such correlated knowledge into a predictor. There  are  many  different techniques  and  tasks related to the use of context in vision, a probabilistic approach of structured prediction in aerial images are methods based on CRFs ~\cite{yao2012describing}.

The advantages provided by an automated aerial image system vary from building monitoring or reconstruction, map creation etc. For a long time hand - crafted features along with simple classifiers provided the best performance for such systems. Some simple methods only use different cues extracted from the image such as color bands, gradients, histograms, geometric features and so on, and based on only one type of information or on a combination of them, one can detect areas in the image  that  contain  the  object  of  interest.  Therefore,  these  methods  focus  on  extracting  multiple features from the input image and detecting the objects  using each feature independently  from each other, and then by using a decision fusion method, the detection results calculated from the previous features are combined. In order to obtain better results for such a difficult problem S{\i}rma{\c{c}}ek et al. ~\cite{sirmaccek2008building} presents a way  of  combining  different image  features,  extracting  several  cues  from  the  main  scene.  The  paper focuses  on  highlighting  the  areas  of  interest,  containing  the  buildings,  using  invariant  color  features, edge  and  shadow  information  and  even  determines  the  shape  of  the  building.  Better  classification results are obtained only after the identification of the difficulties that the input data provides. 

Some  of  the  problems  that  detection  systems  face  are  the  uncontrolled  characteristics  of appearance, cluttered background of a natural scene, and not only that, but in urban areas buildings are generally  dense  and  have  complex  shapes.  In  order  to  overcome  the  difficulty  of  detecting separate houses  from  these  crowded  environments,  an  interesting  method  is  proposed  in ~\cite{sirmacek2011probabilistic}. A probabilistic model is obtained based on the extraction of different feature vectors. In this case the algorithm was tested  on  data  collected  from  different  sources,  images  retrieved  from  two  different  sensors  which provide different spatial resolutions and gives the dataset diversity, namely the test images were very high resolution panchromatic aerial crops and Ikonos satellite images. The main idea of this method is that all extracted local feature vectors are used as observations of a probability density function which must be estimated. What this means is that each and every one of the buildings to be detected in the image are modelled as joint random variables. The location of a building, if it exists in an image, is given by the result of the estimation of the probability function.

One  successful  approach  in  semantic  segmentation,  known  as \textit{autocontext} ~\cite{tu2010auto}, uses classifier outputs from one level of image interpretation as contextual inputs to a higher  level  of  abstraction.  Context  could  be  understood  in  many  forms,  going  from  reasoning  about objects against the global scene ~\cite{torralba2010using} to looking at more precise spatial and temporal relationships and interactions between different object categories ~\cite{leordeanu2014thoughts}. One relevant example is the work of Choi et. al ~\cite{choi2010exploiting} that combines both spatial relations to other objects as well as global scene context.

It is not yet known what is the best way to combine object relationships and global information for contextual reasoning. Deep neural networks are an interesting choice for modeling context, as they process information from one level of abstraction to the next. They use single, discrete neurons, which combined with different ways of pooling could model ”detections” of deferent features, object parts or even  whole  concepts,  at  different  levels  of  abstraction.  Thus  they relate  to  methods  using  object detectors for extra contextual cues. By using many such neurons, with soft responses over potentially large fields, they could also model global image statistics - connecting to literature using whole image contextual features. By reasoning in a hierarchical manner they also offer the possibility of integrating information from one level as contextual input to the next, relating to approaches using autocontext. Therefore, deep nets seem to offer the right environment for designing effective architectures for using and studying visual context. Their recent success in computer vision on various tasks ~\cite{krizhevsky2012imagenet,lecun1989backpropagation,simonyan2014very,girshick2015fast,szegedy2015going,zhang2015improving}
encouraged researchers to start testing different approaches for using context in conjunction with CNNs. 

An even more interesting and novel approach is presented in ~\cite{saito2015building} in which no more hand - crafted features are needed, because of the processing power of deep layered Convolutional Neural Networks (CNN). These deep architectures are used to learn mapping from raw pixel values in aerial imagery to three object labels (buildings, roads, and others). The method discussed in ~\cite{saito2015building} also uses a patch - based semantic  segmentation  approach  as  the  algorithm  proposed  in  the  current  paper.  The  large aerial imagery  is  firstly  divided  into  smaller  squared  patches  of  the  same  size,  and  then  the  CNN  is  trained using the raw unprocessed patches. The novelty of the deep learning system is that the output is based not only on one type of labels, but is able to predict road labels and building labels at the same time. These labels are correlated with each other in urban areas, highlighting the trade-off between road and building at a single pixel with the scope of reducing the confusion between them and improving the performance of prediction. The basic architecture of the CNN used, comprises of stacked convolutional layers and spatial pooling layers often  followed  by one or more fully connected layers. The feature extraction over the input image  is done  in  the  convolutional  layer  which  contains  several  convolution  filters  and  next  a  pooling  layer applies  sub - sampling  to  the  output  of  the  next  lower  layer  for  achieving  translational  invariance. 

Such systems, combining context with deep networks, were proposed for action classification ~\cite{gkioxari2015contextual}, segmentation  by  modeling  CRFs  \cite{zheng2015conditional}  with  recurrent  networks  and  object  detection  by  training contextual networks over nearby bounding box regions ~\cite{zhu2015segdeepm,gidaris2015object}. Other recent work models person context in order to improve detection of objects that are used by or related to people ~\cite{gupta2015exploring}. Another recent  architecture is designed for integrating local and holistic information  for  human  pose  estimation  ~\cite{fan2015combining}.  Note  that  research  in  using  visual  context  for  object detection is also limited by current image datasets, such as PASCAL VOC Dataset ~\cite{everingham2010pascal}, in which objects occupy a large part of the image. 

Different from ~\cite{zhu2015segdeepm,gidaris2015object} our proposed deep architecture is based on a dual - stream network, each pathway having its own different architecture, centered on the object  but looking over different image areas: one considering local information and the other taking into account a much larger region. As we show in our experiments, the two pathways learn by themselves to process the object and its surroundings in two complementary ways, one for finer shape segmentation and the other for reasoning about the larger context. 

Different from previous work, we study context in the domain of aerial imagery, where objects are  relatively  small  and  it  is  easy  to include  larger  areas  as  input.  In  aerial  imagery  most  traditional approaches  are  based  on  multiple  cues  extracted  from  the  image  such  as  color  bands,  gradients, histograms or certain geometric features. Objects  are  first  detected  using  each feature  independently  and  then, by  applying  a  decision fusion method ~\cite{senaras2013building}, the results from previous features are combined. Other work selects the most discriminative features for  semantic  classification  in  aerial  imagery  using  boosting  ~\cite{tokarczyk2015features}.  There  is  also  recent  work  ~\cite{mattyus2015enhancing}  that combines satellite aerial images available online with ground truth labels from Open Street Map (OSM) for  learning  to  enhance  road  maps.  Authors  use  some  weak context  features  based on  differences  in mean  pixels  intensities  between  the  road  area  and  its  background,  within  a  Markov Random Field formulation. Very few approaches in aerial image analysis use CNNs, with improved results ~\cite{saito2015building,MnihThesis}.

In an idealistic scenario, the contributions of using the information surrounding the object does not seem necessary for its recognition. However, when dealing with real - world scenes, where local appearance information of the object is often degraded due to occlusions, illumination, shadows and distance, leading to poor resolution, the role of context is enhanced. When taking into account the object's environment, recognition of the object becomes reliable. Frequently, various object categories may reside in a scene, and the relations among these objects provide complementary contextual cues that help in the recognition task. The idea of local - global information complementarity has been discussed in \cite{torralba2003contextual} and once again enforced by our work.

Our main contribution over the prior work is to show that contextual information is important for accurate object recognition in aerial images and also provide a novel dual - stream architecture, based on deep  convolutional  neural  networks,  which  learns  in  parallel  to  recognize  objects  from  two complementary views, one from the local level of object appearance and the other from the contextual level of the scene.
\chapter{Problem description}
\label{chapter:problem-description}

In this chapter we formulate our semantic segmentation problem and motivate our design and implementation choices. We start by using an state-of-the-art approach in aerial image segmentation and then integrate and motivate the importance of context for this task. We treat this problem from two different perspectives, based on the type of knowledge, with the help of deep CNNs and motivate our proposed network.

\section{Preliminary work and intuition}
\label{sec:intuition}

Local appearance is often not sufficiently informative for segmentation in low - resolution aerial images. Such an example is presented in \labelindexref{Figure}{img:fig_intuition} \textbf{A}, where we can observe two local patches and their larger scene context. By looking only at the smaller patches, it appears that local appearance is not sufficient for confidently recognizing the presence and the shape of a house. In fact, from the local patch alone, the example on the left seems to be more likely to belong to a house than the one on the right since its alignment of features resembles a rectangular structure. When we consider the larger contextual neighbourhood, the house roof is more clearly perceived in the second case, in which the larger residential area contributes in an important way to the local perception. Geometric grouping cues such as agreements of houses’ orientations and similar appearances in the  larger  residential  area  increase the  chance  that  we  are  indeed  looking  at  a  house  and  also  help in perceiving its shape better.

The larger context could provide vital information even for highly localized tasks such as fine object segmentation: the exact shape of the house in the example on the right is better perceived when looking at the larger residential area, which contains other houses of similar shapes and orientations. Thus, local structure could be better interpreted in the context of larger scene.

When looking at the smaller patches and their larger scene context, it appears that the local appearance is not sufficient for confidently recognizing the presence and shape of the house. When looking at the larger contextual neighbourhood we see that this area contributes to the local perception not only for determining the presence or absence of a certain object but it’s also important for a more accurate perception of shape.
In the case on the left, the contextual alignment of the diagonals in the larger region  of  grass  lowers the  possibility that  we  are  indeed  looking  at  a  house.  We argue that contextual influences are not only important for determining the presence or absence of a certain object class, but are also important for a more accurate perception of shape.

\fig[scale=0.39]{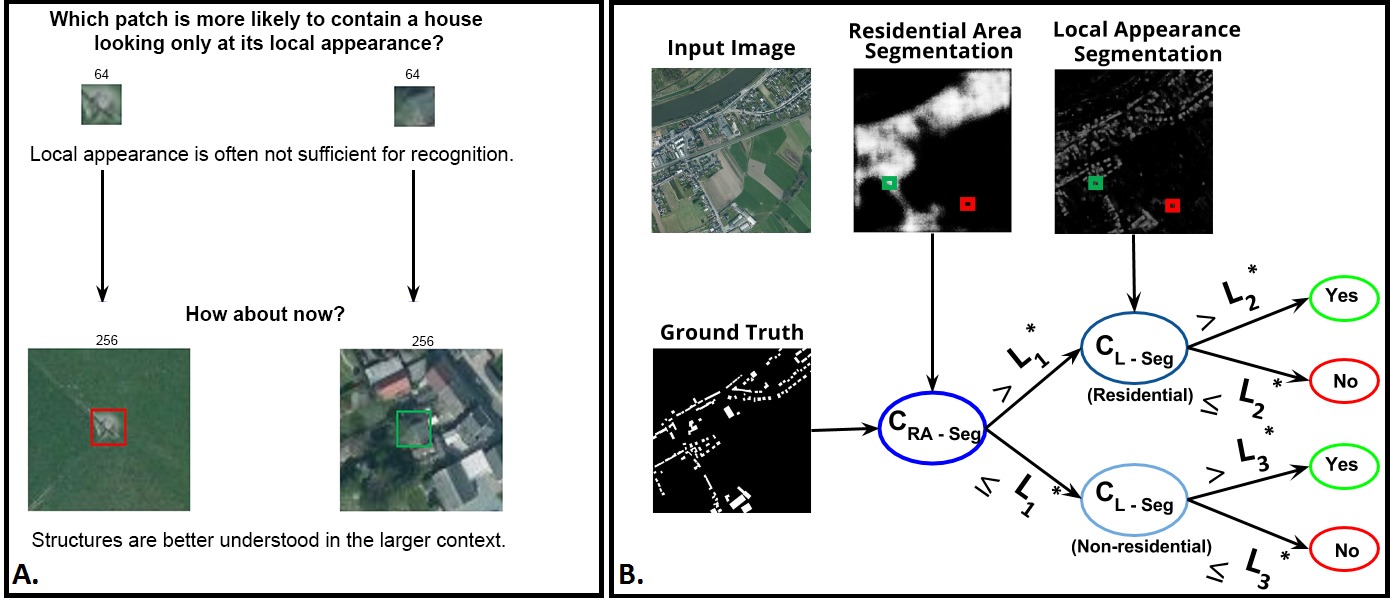}{img:fig_intuition}{\textbf{A.} Comparison between local and global appearance. \textbf{B.} Combining residential segmentation results with the local appearance segmentation results using a tree - like structure.}

We study the role of the local and global information separately and focus on the problem of building segmentation. We consider the problem of finding the shapes of buildings in an aerial image. We treat the task from two perspectives, considering both their local appearance as well as information from the larger scene containing them. We are interested to study the role of context on this task, as buildings have various shapes and appearances and are representative for most aerial images. We employ two models based on deep CNNs that solve this problem separately and then jointly.

\section{Locally - informed VGG-Net}
\label{sec:locally_informed}

Current state-of-the-art approaches use deep convolutional neural networks for the segmentation task in aerial images. As previously stated, Shunta Saito  and  Yoshimitsu  Aoki developed such an architecture and published their solution at the beginning of 2015 ~\cite{saito2015building}. Their method uses a patch - based semantic segmentation approach. So firstly the large aerial image is divided into small patches further used in the training of the CNN. They employ this method in order to create a robust feature extractor from raw pixel values and at the same time learn classifiers for the building recognition task. The model is  trained  with a  huge  amount  of patches  of non - centered individual  buildings. The labeling process is done pixel - wise, for each pixel of the RGB patch  there  is  a  corresponding  binary  label  in  the  ground truth map, 1  in  case  the  pixel  belongs  to  a building and 0 otherwise. The model is trained such that at prediction, it outputs a probability for each pixel, highlighting the areas in which the classifier has a high confidence in the building prediction. Their network receives as input patches of size 64 x 64 pixels and outputs 16 x 16 patches of pixel - wise labels.

We used a similar approach to train our networks. The architecture of the local appearance model was inspired from the state-of-the-art object detection architecture VGG-Net ~\cite{simonyan2014very}. The  main  contribution  of  the paper was  showing  that  the  depth  of  such  models  is  a  critical component for good performance. The authors varied the number of stacked layers in order to determine the optimal number of layers that provide the best detection performance. The  entire  neural  network  is  composed  of  convolutional  layers  that  perform convolutions using 3 x 3 kernels with stride 1 and pad 1, followed by pooling layers that perform 2 x 2 max pooling  with  stride  2  and  no  padding,  and  then  continued  with  rectifier  units.  Despite  slightly weaker  classification  performance, the  VGG-Net  features  outperform  those  of other  state-of-the-art methods in multiple transfer learning tasks. Its main downside is that it is more expensive to evaluate since  it  uses  a  lot  more  memory  and  it  is  a  highly  complex  architecture  given  the  total  number of parameters (around 144 millions). 

\fig[scale=0.3]{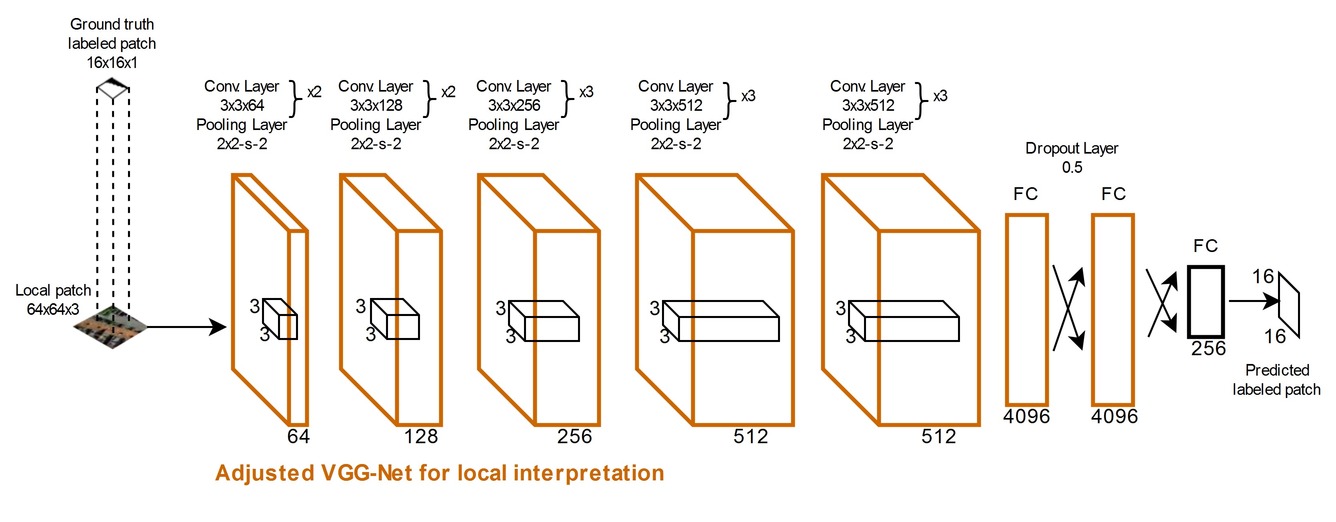}{img:fig_locally}{VGG-Net architecture trained using only local information.}

At test time the image is divided into a disjoint set of patches, on a grid, and each patch is classified independently. The end result becomes a segmentation of the entire image, with white areas belonging to building pixels. The input to the network is a 64 x 64 patch that, in the case of houses, often contains little surrounding background information. This network is thus trained to detect and segment houses (output their exact shapes) using mostly local information. To avoid any misinterpretation, the variant VGG - Net implemented in this paper will be referred as L - Seg (individual building segmentation using only local information) model and its main architecture is shown in \labelindexref{Figure}{img:fig_locally}.

We train this network on different datasets and report the results in \labelindexref{Chapter}{chapter:experiments}.

\section{Residential area segmentation}
\label{sec:residential_segmentation}

In order to study the role of the larger context, we employ a wider (with larger filters and input) but shallower architecture based on AlexNet ~\cite{krizhevsky2012imagenet}, which takes as input a 256 x 256 image patch (16 times larger in area than the input to L - Seg). This model is not trained for accurate shape prediction, but only to output a single binary variable - whether the input patch belongs to a residential area or not. Since the goal of this network is to classify patches of residential versus non - residential areas, there is no need to use the same complexity and dimension of the previously detailed model, since in  this  case  the classification and  label  is  given  per  patch  not  per  pixel,  the  complexity  of  the  problem reduces drastically.

The  scope  in this case  is to train  a  two - class patch - based classifier.  Depending on the  area,  rural  or urban, the density of the houses varies. Sparse houses are not considered residential areas, but they also  remain of interest in the detection problem. The goal of this model is to roughly segment the regions of interest,  the  regions  where  there  is  a  high  probability  of  buildings  encounters. In this  way,  various classification mistakes, such as false positives detected from the previous L - Seg model, are eliminated. The regression  of  the  previously  obtained  results  is undesirable,  therefore  it  was  necessary  to  divide  the residential areas in 3 main categories depending on the number houses present in the patch:

\begin{itemize}  
\item \textbf{Category I} : patches containing 1 – 5 buildings 
\item \textbf{Category II}: patches containing 6 – 15 buildings 
\item \textbf{Category III}: patches containing 16 – 30 buildings 
\end{itemize}

For each of the previously mentioned categories, a different classifier was trained in the same manner without tuning the learning parameters, from a model to another. Special attention was paid in the  patch  extraction  process. The  size of  a  patch is  determined  by  the  maximum  value between  the  width  and  the  height  of  a  bounded  box  containing  the  house. All  the  patches  have the houses centered and the area surrounding the object of interest, denoted as context, is 6 types bigger than the dimension of the house. We will refer to this model as residential area segmentation model or simply RA - Seg.

The training set in this case contains a total of 25600 patches, with a positive : negative samples ratio  of 1 : 7.  A  validation set is  also  provided  containing  10\%  of  number  of  patches  contained  in  the training set, meaning a total of 2560 patches in total. Positive patches are considered patches containing residential  areas  from  one  of the categories  considered,  whilst  negative  samples  are  patches  without any buildings in the scene, such as forest, meadow, river, roads etc. The training characteristics from the previous model were preserved as much as possible for the all the residential category models. Although the number of training epochs for each model varies, the results are reported with reference to the model trained at 10 epochs. The results of in minimizing the objective function are shown in \labelindexref{Figure}{img:fig_objective}. 

\fig[scale=0.5]{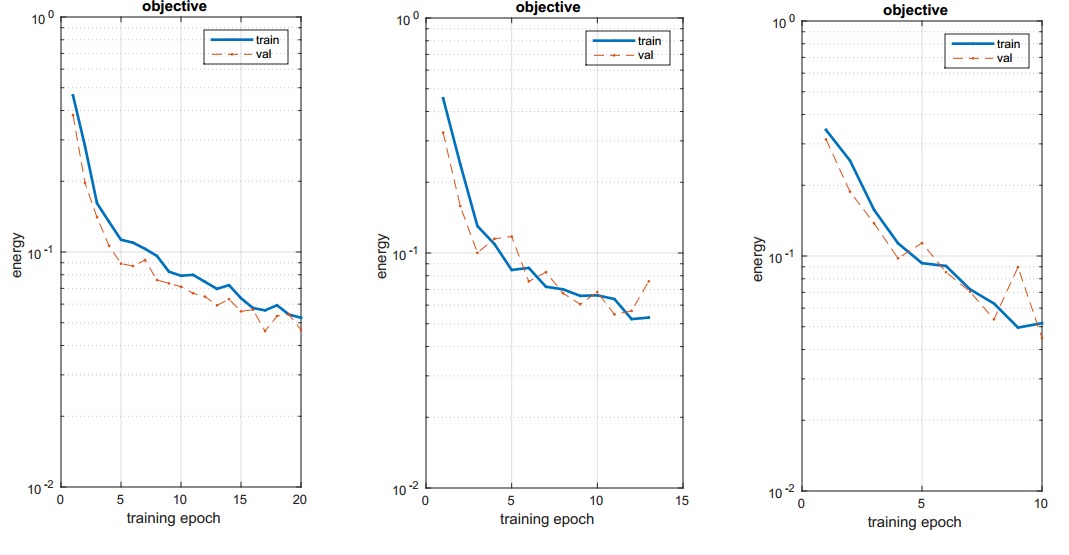}{img:fig_objective}{Visualization of the objective minimization for the residential area classification models. From left to 
right in order, Category I, Category II, and Category III Loss.}

\begin{table}
\begin{center}
{\caption{F-measure scores for RA - Seg models trained on different house density patches, from low density (Cat. I), to medium density (Cat. II) and high density (Cat. III)}\label{tab:categories}}
\begin{tabular}{l*{6}{c}r}
\hline
\rule{0pt}{10pt}
Threshold & 0.1 & 0.3 & 0.5 & 0.7 & 0.9 \\
\hline
F-measure (\textit{Category I}) & 0.3145 & 0.3135 & 0.3079 & 0.3038 & 0.2987 \\
\hline
F-measure (\textit{Category II}) & 0.3160 & 0.3129 & 0.3074 & 0.3034 & 0.2983  \\
\hline
F-measure (\textit{Category III}) & 0.3165 & 0.3138 & 0.3085 & 0.3042 & 0.3000  \\
\hline
\end{tabular}
\end{center}
\end{table}

Since the all of the trained models offer similar performance (see \labelindexref{Table}{tab:categories}, the optimal threshold that gives 
the  best  F - measure  over  all  the testing  images for  the  RA - Seg  model  is  computed  only  for  the  third 
category (high density regions). For the optimal threshold of 0.11 the value of F - measure for the third model is 0.3167. The performance for the residential area segmentation models could not be correctly computed since residential area labels were not provided, therefore we decided to use the same ground truth for our evaluations as the one used for the local segmentation of individual buildings, with precise shape.

Our goal was to see whether the two models, trained completely separately on two different tasks, one for accurate shape segmentation and the other for simple binary classification could be later combined for improved performance, since they were trained on different types of information.

\fig[scale=0.9]{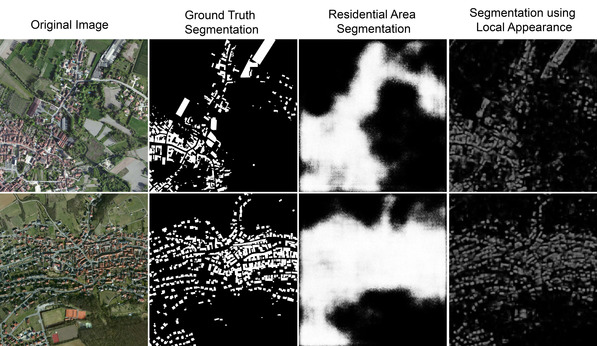}{img:fig_ra_results}{Qualitative results of our residential area segmenter trained with high density housing patches compared to a segmenter trained only with local appearance information.}

\section{Initial experiments}
\label{sec:initial-experiments}

Based on the previously presented qualitative results, some important observations are worth mentioning. Firstly, it is noticeable that the residential area segmentation performs better in terms of removing unwanted output, such as the mass of false positive responses to which the L - Seg model is so sensitive. This promising result can be justified by the fact that these models were trained on different types of data, L - Seg model uses only local information whilst RA – Seg was trained not only with local information, but also extra information received by the neighbouring area. The information available in the scene around objects as well as relations among them is known to provide complementary contextual cues for recognition. Such cues are likely to be particularly helpful when the appearance of the object lacks discriminative cues due to low image resolution, poor lighting or other factors, which is the case of the challenging recognition problem discussed in this thesis.

\fig[scale=0.32]{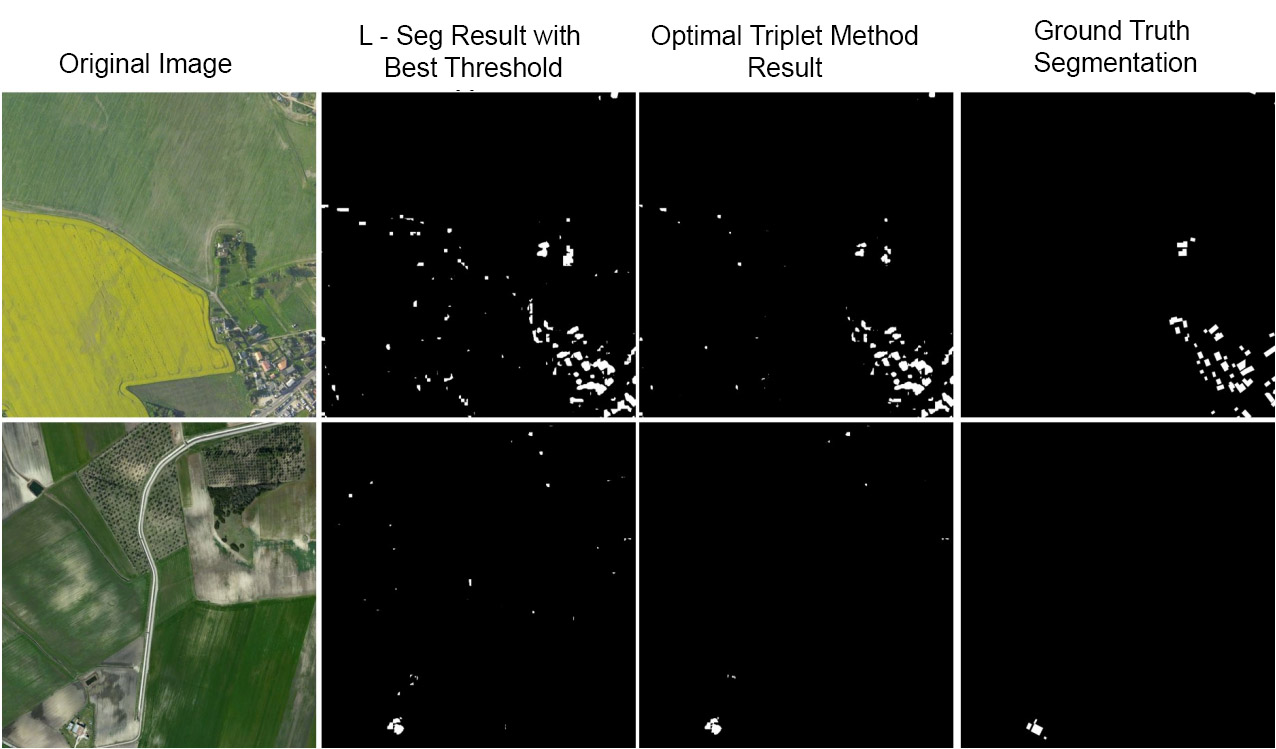}{img:fig_initial_experiments}{Buildings segmentation results using the combined tree model (third column) versus the local segmentation (L - Seg) alone (second column) without the residential area detector (RA - Seg).}

The sequence of steps necessary for combining both outputs from the previous trained networks is further detailed. The scope of this method is to determine an optimal set of thresholds capable of improving the recognition performance of just the simple L - Seg prediction. This is the first step of the process. In the same manner the other two thresholds are iteratively changed in small steps, improving the overall recognition rate of the testing set. The method loops and iteratively tries various combinations of thresholds until no more improvements can be made. It receives as input a set of previously unseen images, consisting of three types of maps, namely the ground truth maps, the results of the RA – Seg model and the results of the L – Seg model and also a set of threshold values (0.01 : 0.01 : 1, the set used in our experiments). Based on the ground truth maps, each image is evaluated iteratively using all the values in the threshold vector. The selected threshold is applied on the results of both models individually and the mean F1 – Score over all images is computed. The maximum F1 – Score over all thresholds determined the best threshold for the models. The best RA – Seg threshold is then applied to all RA – Seg maps. We name this threshold $L_{1}$. The maps of the L – Seg model are divided into two main regions based on the thresholded RA – Seg maps: a map highlighting the non – residential area of the L – Seg map, respectively the residential area of the map. The best L – Seg threshold, called $L_{2}$ in our experiments, is then applied to the residential area of the map, whilst the optimal threshold for the non – residential area, $L_{3}$ is determined by applying all the values of the threshold vector and determining the one that offers the best pixel – wise recognition performance, based on the computed F1 – Score. The algorithm iterates through triplets combinations, until no improvement can be made and returns a triplet of the optimal thresholds. The sequence of steps made by our method for computing the optimal triplet ($L_{1}$, $L_{2}$, $L_{3}$) are presented in \labelindexref{Table}{tab:tree_results}.

We start our algorithm with fixed values of $L_{1} = 0.11$ and $L_{2} = 0.45$ for which we obtain the best F - measure for the residential area segmenter, 0.3167 (RA - Seg), respectively 0.5979 (L - Seg).

\begin{table}
\begin{center}
{\caption{Iterations of our algorithm for computing the optimal triplet of thresholds for the tree - like classifier.}\label{tab:tree_results}}
\begin{tabular}{l*{6}{c}r}
\hline
\rule{0pt}{10pt}
$L_{1}$ Fixed, $L_{2}$ Fixed, $L_{3}$ Iteration  & $L_{1} = 0.11$,  $L_{2} = 0.45$,  $L_{3} = 0.60$ (\textbf{0.6035}) \\
\hline
$L_{1}$ Iteration, $L_{2}$ Fixed, $L_{3}$ Fixed & $L_{1} = 0.08$,  $L_{2} = 0.45$,  $L_{3} = 0.60$ (\textbf{0.6047}) \\
\hline
$L_{1}$ Fixed, $L_{2}$ Iteration, $L_{3}$ Fixed & $L_{1} = 0.08$,  $L_{2} = 0.41$,  $L_{3} = 0.60$ (\textbf{0.6056}) \\
\hline
$L_{1}$ Fixed, $L_{2}$ Fixed, $L_{3}$ Iteration & $L_{1} = 0.08$,  $L_{2} = 0.41$,  $L_{3} = 0.62$ (\textbf{0.6057}) \\
\hline
$L_{1}$ Fixed, $L_{2}$ Iteration, $L_{3}$ Fixed & $L_{1} = 0.08$,  $L_{2} = 0.425$,  $L_{3} = 0.62$ (\textbf{0.6058}) \\
\hline
$L_{1}$ Iteration, $L_{2}$ Fixed, $L_{3}$ Fixed & $L_{1} = 0.08$,  $L_{2} = 0.425$,  $L_{3} = 0.62$ (\textbf{0.6058}) \\
\hline
\end{tabular}
\end{center}
\end{table}

Our qualitative results \labelindexref{Figure}{img:fig_initial_experiments} are obtained after we applied the optimal triplet of thresholds ($L_{1} = 0.08$, $L_{2} = 0.425$, $L_{3} = 0.62$).

Our initial model for residential area detection (RA - Seg) has poor localization but low false positive rate within a larger neighbourhood. RA - Seg can be effectively combined, in a simple classification tree, with the local semantic segmentation model (L - Seg), which has higher localization accuracy but
also a relatively high false positive rate. In \labelindexref{Figure}{img:fig_intuition} \textbf{B} we see how the output from RA - Seg can be used in order to filter out the houses hallucinated by the local L - Seg model.

\section{Discussion}
\label{sec:discussion}

We found that the two models can be effectively joined into a classifier tree, with the residential area  classifier  (RA - Seg)  acting  as  a  filter  that reduces  the  false  positive  rate  of  the local  buildings  shape segmenter L - Seg (\labelindexref{Figure}{img:fig_initial_experiments}). While the L - Seg CNN segments disjoint patches on a grid, the RA – Seg classifier gives single labels to those patches. A dense pixel - wise residential area classification could be obtained by interpolation. The tree model is formed and presented in \labelindexref{Figure}{img:fig_intuition} \textbf{B} by putting the RA - Seg classifier at a first node and the L - Seg model at the leaves. Depending on how the first node classifies the patch, the leaves will classify it using different thresholds. Consequently, if a patch is classified as residential by RA - Seg, the segmenter L - Seg will be more likely to detect buildings than otherwise. The  tree  is  controlled  by  the two  models  with  three  different thresholds $L_{1}$, $L_{2}$  and  $L_{3}$. $L_{1}$  is applied  to  the  RA - Seg  classifier,  while  $L_{2}$  and  $L_{3}$  control  the  precision  of  the  L - Seg  leaves.  The  three parameters  are  optimized  in sequence,  until  convergence,  as  follows:  before the  first  iteration,  the thresholds  are  chosen  independently  to  maximize  the  mean F - measure  of  the  two  classifiers.  Then, each threshold is optimized in turn, while the other two are kept fixed. The F - measure is thus improved from 59,8\% to  60,6\% on the European Buildings Dataset (presented in detail in \labelindexref{Table}{tab:tree_results}). Note that these numbers  are relatively  low  compared  to  the  ones from  the  experimental  section presented in \labelindexref{Section}{sec:european_buildings}, because  on  these initial experiments we stopped the training of the CNNs relatively early, before complete convergence in order to validate our intuition. Also,  for  evaluation  we  did  not  use  the  relaxation  of  three  pixels  which  we  applied  later,  in  order  to compare  with  other  methods.  At  this  point  all  we  are interested  in,  is  whether  a  residential  area detector can be combined effectively and in a simple way with a local buildings segmenter. We should also note that the overall quantitative improvement of 0,8\% is an average value over all pixels in the test set. It does not capture the more qualitative benefit of using the RA - Seg classifier, which is able to filter out buildings that are hallucinated by the local segmentation in areas of high texture (as shown in \labelindexref{Figure}{img:fig_initial_experiments}). Since  buildings generally  occupy  only  a  small  fraction of  pixels,  the  overall  average improvement  is significantly less than the improvement in those specific places.

\section{Globally - informed AlexNet}
\label{sec:globally_informed}
 
In order to study how the larger context influences the shape of the object, we employed the same architecture used for the task of residential area segmentation only in this case we trained AlexNet to be able to determine the precise shape of the object, not just for binary classification. This network was trained with 256 x 256 patches centered at the same location as the patches used to train the L - Seg model. \labelindexref{Figure}{img:fig_globally} presents the detailed layers of the trained architecture. The results of this model are reported in \labelindexref{Section}{sec:european_buildings}. 

\fig[scale=0.29]{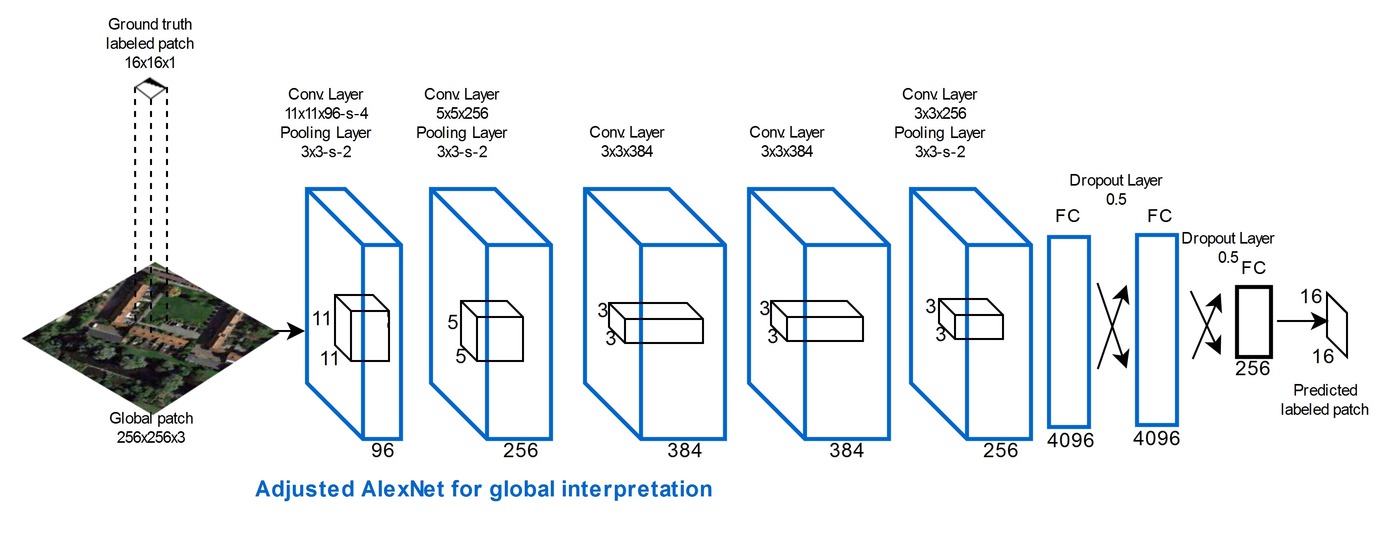}{img:fig_globally}{Alex-Net architecture trained using only global information.}
 
\section{A dual local - global CNN for semantic segmentation}
\label{sec:local_global_cnn} 
 
\subsection{Motivation}
\label{sub-sec:motivation}

We take the intuition and initial tests from the previous section a step further and create an architecture that combines the previous two models into a single local - global deep network, termed LG - Seg (our combined architecture is presented in \labelindexref{Figure}{img:fig_combined}). Our proposed network is formed by modifying and joining two state-of-the-art deep nets, namely VGG - Net used here for local image interpretation ($L - Seg$) and AlexNet - used here for global interpretation of the contextual scene ($G - Seg$). Note that the L - Seg network is deeper but narrower with smaller filter sizes (and smaller input in our case) and it is thus better suited for more detailed local processing. G - Seg network, which is shallower (fewer layers) but wider (larger input and filters), takes into consideration more information at once and it is thus more appropriate for global processing of larger areas. The two pathways meet in the final fully - connected layers, which combine information about object and context into a unified and balanced higher level image interpretation. Our network is trained jointly, end-to-end on various datasets and applied not only to the problem of building segmentation. More details are given in \labelindexref{Chapter}{chapter:experiments}.

The two pathways process information in parallel, taking as input image patches of different sizes. Then,
the superior fully - connected layers of each individual network are concatenated and fed into three different fully - connected layers that learn how to combine local and contextual information
at the level of semantic interpretation. Features at the final layers in each pathway reach a relatively high level of abstraction. Here we expect the object and its context to reach the final decision - this level is the place where bottom - up and top - down reasoning about objects meet in order to resolve conflicts and reinforce agreements. Based on the experiments performed with the simple tree model we want to find whether the two subnets (\labelindexref{Figure}{img:fig_combined}) indeed learn categories at different levels. The local one focuses more on the exact shape of individual buildings and the other classifies larger residential areas with less focus on exact localization and delineation of buildings.

\fig[scale=0.35]{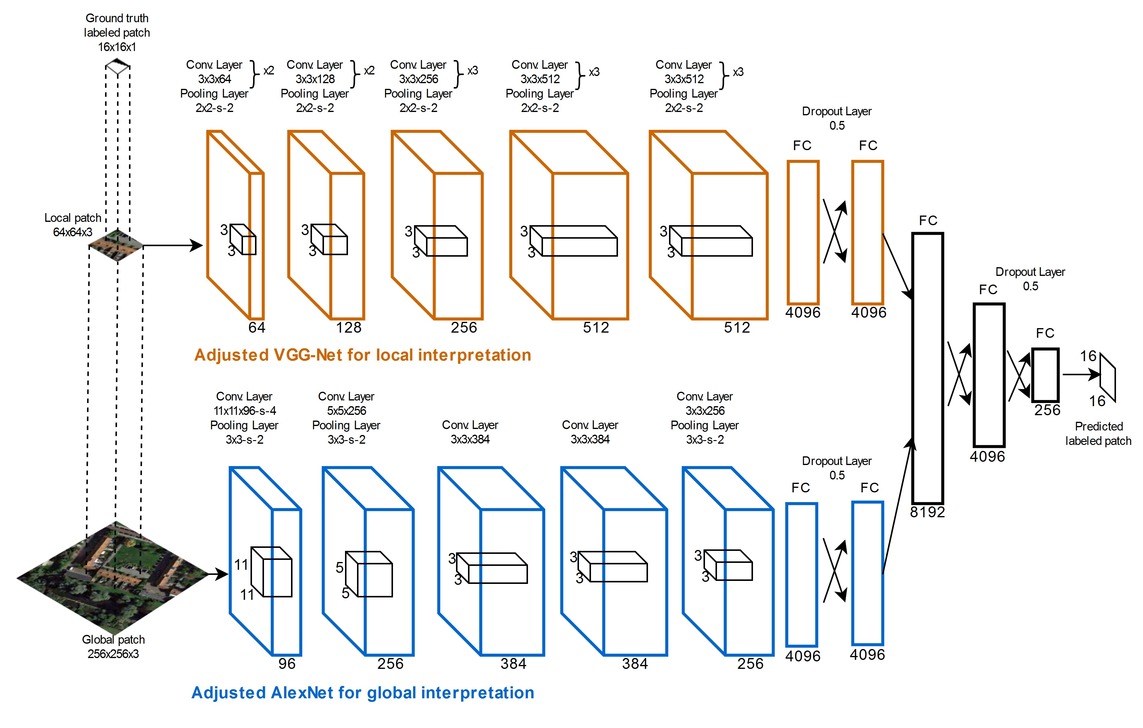}{img:fig_combined}{Our proposed, dual - stream, local - global architecture LG-Net.} 

Why is the dual local - global architecture better than a single very deep one? Our motivation for this dual - stream architecture is straightforward. Besides the computational challenges brought by a deeper net, such as larger memory requirements, quantity of data and time of training, a single net would have functioned more like black box training mechanism, while mixing the local and global information into a single path for processing - and that is exactly what we wanted to avoid here.

Residential areas are a different category, on its own. For example, small green spaces between buildings, side - walks, parking lots or playgrounds for children 
may all be part of the residential area but they are not buildings. However, their existence is indication of the presence of buildings. Residential areas \emph{exist} over large regions and at higher level of semantic abstraction: they are regions where people live and could even form communities, well beyond the idea of houses or buildings.
While a small patch of grass or concrete could be part of a residential region, the same patch of grass or concrete, when present inside a large park or on an important street, should not be seen as part of a residential place. 
This aspect of complementarity between an object, such as a building, and its scene, such as the residential area,
is exactly what we want to study: what are the two pathways learning when initializing the whole network from random weights? Our experiments, presented in \labelindexref{Section}{sec:local_global_complementarity}
confirm this fact: the two networks indeed learn to process information
in complementary ways, one distinguishing more individual houses and buildings and the other focusing on larger residential areas.

We also expect the single combined network trained end-to-end to be able to produce more accurate segmentations
over the simple tree model. Note that the tree model usually does not improve the shape of the segmentation produced by the local network, but only changes 
the recognition confidence, using two different thresholds, over relatively large areas. In the classifier tree case, the residential area network outputs a single label per patch, while in the LG - Seg model they are jointly trained to segment objects - technically this is a good reason why we expect a qualitative improvements in segmentation of objects shape.  
 
\subsection{Problem formulation and learning}
\label{sub-sec:problem-formulation}

We formulate the object segmentation problem in a way that is similar to the one proposed by
Mnih et. all \cite{mnih2010learning}, as a binary labeling task, where all pixels belonging to the object of interest are 1 and all the others are 0. Let 
$\mathbf{I}$ be the satellite aerial image and $\mathbf{M}$ the corresponding ground truth labeled map. The goal is to predict a labeled image $\hat{\mathbf{M}}$ from an input aerial image $\mathbf{I}$, that is to learn $P(M_{ij} | \mathbf{I})$ from data, for any location $p=(i, j)$ in the image.

We train our network to predict a labeled image patch $W(\mathbf{M}, p, w_{m})$, extracted from labeled map $\mathbf{M}$, 
centered at location $p$, of window width $w_{m} = 16$, 
from two aerial image patches $W(\mathbf{I}, p, w_{l})$ and $W(\mathbf{I}, p, w_{g})$, centered at the same location $p$, with a smaller size window width 
$w_{l}=64$ for the local patch and a larger window width $w_{g}=256$ for the global patch.
We want to learn a mapping from raw pixels to pixel labels and use  
a loss function that minimizes the total cross entropy between ground truth patches and predicted label patches. 
For each forward pass during learning, LG - Seg receives as input three types of patches, the 16 x 16 patch from the ground truth map, the local 64 x 64 image patch 
and the global 256 x 256 context patch, centered at the same point and having the same spatial resolution (see \labelindexref{Figure}{img:fig_combined}).

Given a set of \textit{N} examples let $\hat{\mathbf{m}}^{(n)}$ be the predicted label patch for the \textit{$n^{th}$} training case
and $\mathbf{m}^{(n)}$  the ground truth patch. Then our loss function $L$ is:

\begin{equation}
\label{eq:learning}
L=-\sum\limits_{n=1}^N \sum\limits_{p=1}^{w_{m}^{2}} (m_{p}^{(n)} \log \hat{m}_{p}^{(n)} + (1 - m_{p}^{(n)}) \log (1 - \hat{m}_{p}^{(n)}))
\label{distutv}
\end{equation}

\subsection{Training details}
\label{sub-sec:combined_training_details}
The minimization of this loss is solved using stochastic gradient descent with mini - batches of size 10 for all the training sets that we used. Momentum was set to 0.9, starting with a learning rate of 0.0001 and $L_{2}$ weight decay of 0.0005. We initialize the weights using the Xavier algorithm, in order to deal with the problem of vanishing or exploding gradient
during learning in deep networks. This method automatically determines the scale of the initial weights based on the number
of input and output neurons, in order to keep the weights within a reasonable range.
All our learning and testing was ran on GPU GeForce GTX 970, with 4GB memory and 1664 CUDA cores. 
Our models were implemented, trained and tested using Caffe~\cite{jia2014caffe}.

\chapter{Experimental analysis}
\label{chapter:experiments}

In this chapter we test our theory on different segmentation tasks. Firstly, we perform experiments  on  finding  buildings  on  two  large - scale datasets  from  different regions in the world: USA and Western Europe. And report state-of-the-art results on one publicly available dataset. We then used our method for the purpose of roads segmentation on Romanian territories. These datasets vary greatly in terms of quality and content. We extend our application to other classes and train different networks for segmentation of meadows, forest and water.
 
\section{Evaluating our models} 
\label{sec:evaluation}

For the evaluation of each model, we used a qualitative measure as well as a quantitative one. The model is trained such that at a forward - pass through the network it outputs  probability for each pixel, highlighting the areas in which the classifier has a high confidence in the building prediction. The qualitative metric of evaluation involves a visual representation of the detected object and an accurate perception of its shape. 

In the case of quantitative evaluation of the models, the most frequently used metric for the evaluation of detection systems is the precision - recall curve. In the remote sensing literature, precision and recall are also known as $correctness$ and $completeness$. It is common practice to evaluate high resolution data detectors using a relaxed version of these measures ~\cite{wiedemann1998empirical}. The relaxed version of correctness represents the fraction of predicted building pixels that are within $\rho$ pixels of a true building pixel, whilst the relaxed completeness represents the fraction of true building pixels that are within $\rho$ pixels of a predicted building pixel.

\begin{equation}
\label{eq:precision}
Correctness =\frac{true \, positives}{true \, positives + false \, positives}
\end{equation}

\begin{equation}
\label{eq:recall}
Completeness =\frac{true \, positives}{true \, positives + false \, negatives} 
\end{equation}

\section{Detection of Massachusetts Buildings}
\label{sec:mass_buildings}

We start by experimenting with the relatively recent Massachusetts Buildings Dataset~\cite{MnihThesis}.
It consists of 151 high quality aerial RGB images of the Boston area. They are of size 1500 x 1500, at resolution $1 \, m^{2} / pixel$, and represent
mostly urban and suburban areas, containing larger buildings, individual houses and sometimes even garages.  
The entire dataset covers roughly $340 \, km^{2}$. 
It is randomly divided in a set of 137 images used for training, 4  used for the validation of the model and 10 images for testing. 
We extracted approximately 700K patches from the training images and trained our model over 13 epochs for about 4 days on the GeForce GTX 970.

\fig[scale=0.82]{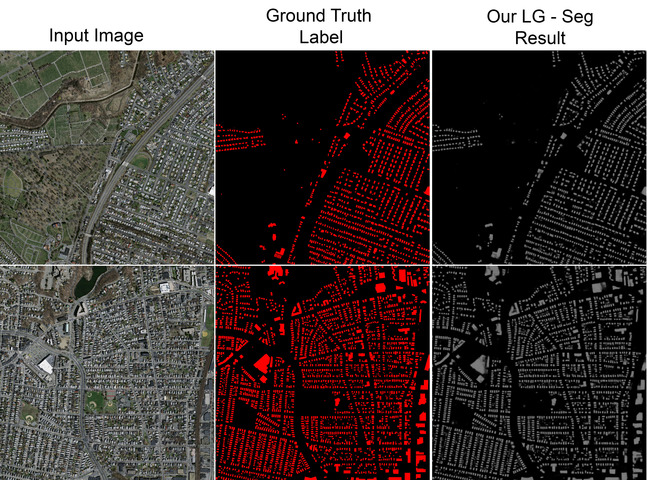}{img:fig_mass_results}{Qualitative buildings detection results on the Massachusetts Buildings Dataset.} 

For computing the maximum mean F - measure over the testing set we applied the same relaxation of 3 pixels used by the competitors:
for a given classification threshold, a positively classified pixel is considered correct if it is within 3 pixels from any positive
pixel in the ground truth map. This relaxation provides a more realistic evaluation, 
as borders of buildings in ground truth are often a few pixels off.

In our buildings  detection  results  on  the  Massachusetts Buildings Dataset (\labelindexref{Figure}{img:fig_mass_results}) we observe the high level of regularity of buildings and roads, which look
very similar to each other. This permits the deep nets to learn perfectly and almost match human performance.

\begin{table}
\begin{center}
\caption{\textbf{Results on Massachusetts Buildings Dataset.}}
\label{tab:Mass_Results}
\begin{tabular}{lccc}
\hline
\rule{0pt}{10pt}
 Method&Mnih et al. ~\cite{mnih2013machine}&\multicolumn{1}{c}{Saito et at. ~\cite{saito2015building}}&\multicolumn{1}{c}{Ours}
\\
\hline
\\[-6pt]
F-measure &0.9211&0.9230& \textbf{0.9423} \\
\hline
\end{tabular}
\end{center}
\end{table}

The significant difference between our approach and the previous state-of-the-art on the Massachusetts dataset in the high 90\% F - measure regime is worth mentioning. While the improvement between 2013 and 2015 was less than 0.5\%, we brought a significant 2\% improvement, from 92.3\% to 94.2\%, thus reducing the error rate by ~25\%.  The high detection rate, above 90 percent is due to data quality. Also, the database does not offer a lot of variation, notice the high level of regularity of buildings and roads, which look very similar to each other. Therefore we have collected our own dataset from areas randomly distributed all over Europe. 

\section{Detection of European Buildings}
\label{sec:european_buildings}

Next we tested on the European Buildings Dataset, which we collected from Western European urban and suburban areas. They contain
a lot more variation than in US, in terms of general urban structure and roads, architecture style, layout of green spaces versus residential areas and geography.
We have gathered 259 RGB satellite images from Google and Bing maps, of size 1550 x 1600 pixels, with spatial  resolution of about $0.8 \, m^{2} / pixel$, with locations picked randomly from different Western European countries.
Covering a larger total area of $348.5 \, km^{2}$ of urban and rural regions spread across Europe,
these images also had a lot more variation in illumination as compared to those from Boston.
We randomly selected 144 images for training ($198.2 \, km^{2}$), 10 for validation ($21.3 \, km^{2}$) and 100 for testing ($129 \, km^{2}$).
The ground truth labeled map for each individual image was generated using data from the Open Street Map (OSM) project. We automatically aligned the satellite images with their corresponding maps from OSM, which has manually annotated buildings. For training we extracted about 1 million patches.

\fig[scale=0.34]{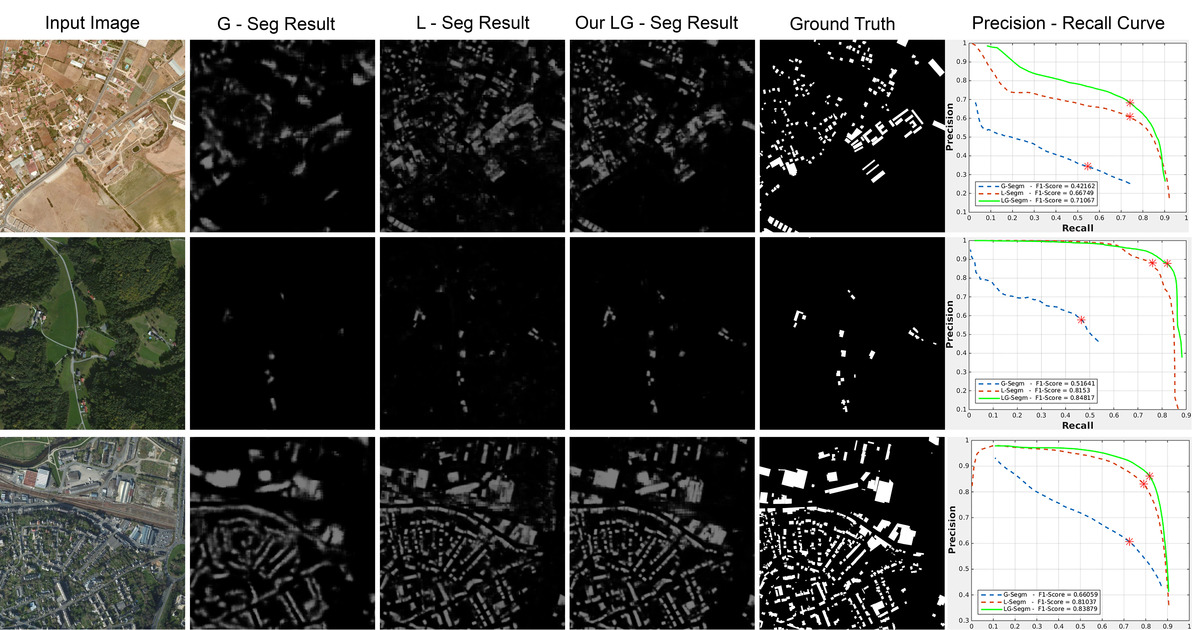}{img:fig_european_results}{Results comparison between the local L - Seg, global G - Seg and local - global LG - Seg architectures.} 

We tested three models (results are presented in \labelindexref{Table}{tab:European_results} and \labelindexref{Figure}{img:fig_european_results}): 
our full LG - Seg net and models formed by keeping only one pathway, G - Seg with the adjusted AlexNet only and
L - Seg formed by the adjusted VGG - Net only. We wanted to test the capabilities of each separately and study the potential
advantage of combining them into a single LG - Seg. All models were trained until complete convergence of the loss, with the G - Seg model taking $34$ 
epochs, L - Seg model $23$ epochs and LG - Seg converging the fastest, in only $12$ epochs. Training time varied between 3 to 6 days on our
GeForce GTX 970. 

Note in \labelindexref{Table}{tab:European_results} that LG - Seg is superior, with over $1.5\%$ improvement in F - measure, on average, over L - Seg. The improvement is significant especially in regions of low residential density where the local model tends to hallucinate buildings. Note in \labelindexref{Figure}{img:fig_european_results} that G - Seg does poorly by itself as it cannot capture fine segmentation details, but it becomes valuable, as a scene processing pathway, within the LG - Seg framework. By reasoning over a larger
area LG - Seg is able to remove false positives (see column four for results) and is also able to produce more accurate building shapes. We stress out that the qualitative difference between the local - global approach and the single deep net is clearly visible on the output map in non - residential areas where the single net hallucinates houses. As these structures are very small, the false positives do not affect the average F - measure by a large value, numerically. Thus the 1.5 - 2\% quality difference is significant in aerial imagery where the positive structures are relatively very small.

\begin{table}
\begin{center}
{\caption{\textbf{Results of our trained models on the European Buildings Dataset.}}\label{tab:European_results}}
\begin{tabular}{lccc}
\hline
\rule{0pt}{10pt}
 Method&{G-Seg}&\multicolumn{1}{c}{L-Seg}&\multicolumn{1}{c}{LG-Seg}
\\
\hline
\\[-6pt]
F-measure &0.6271&0.8266& \textbf{0.8420}\\
\hline
\end{tabular}
\end{center}
\end{table}

\section{Detection of Romanian Roads}
\label{sec:romanian_roads}

We have collected aerial images of two Romanian cities, Cluj and Timisoara, of size 600 x 600 and resolution $1 \, m^{2} / pixel$ and automatically aligned them with OSM road maps to obtain the ground truth labels. 
For Cluj we have 3177 images covering an area of about $70 \, km^{2}$, and for Timisoara 4027 images for an area of $72 \, km^{2}$. Images have significant spatial overlap,
such that there is one image for each road intersection (as estimated from OSM).
For this dataset we trained our model on the task of road detection, as roads are the only category represented relatively well in OSM
over these Romanian regions. We used Timisoara images for training our LG - Seg model and Cluj images for testing.

This dataset provides a much more challenging task due to limitations and variations in the data for the road detection problem.
Different from the other image datasets, this one is of significantly lower quality, with large variations in the road structure, their type, width and length. Moreover, often the roads are completely occluded by trees and the OSM road maps do not match correctly what is seen in the image (see examples in \labelindexref{Figure}{img:fig_roads_results}).
Also note that Timisoara and Cluj have different urban styles, which
brings an extra degree of difficulty for learning and generalization.
For these many reasons, on this dataset, the problem of recognition is tremendously difficult and pushes the limits of deep learning to a next level, as reflected by the significantly lower performance.

\fig[scale=0.36]{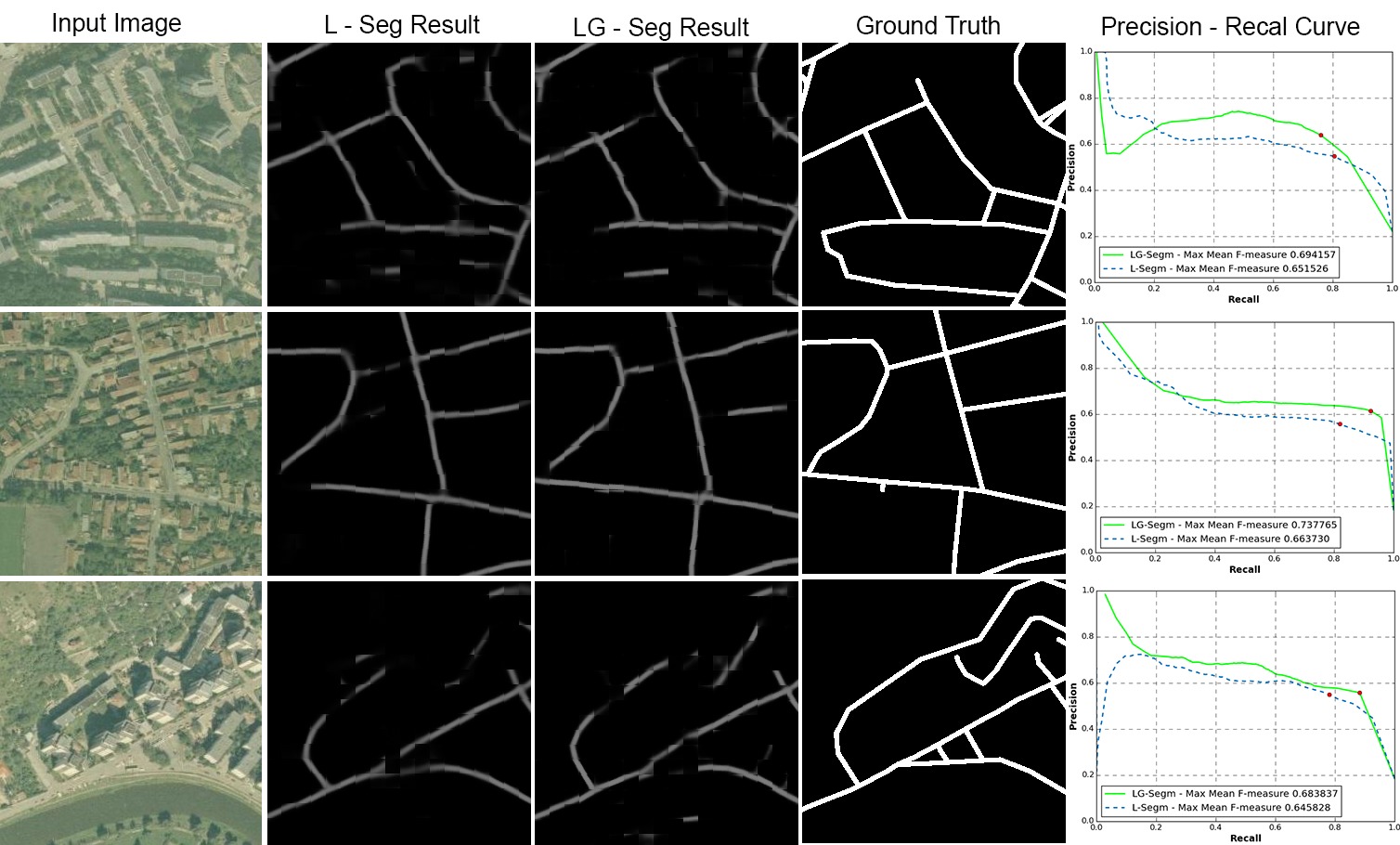}{img:fig_roads_results}{Example results on Romanian roads. Note how difficult the  task is on these images, posing a real challenge even for humans.} 

In Table \labelindexref{Table}{tab:Romanian_roads} we present results and comparisons between the LG - Seg and L - Seg models
on the Romanian Roads Dataset. Again, both quantitatively and qualitatively the LG - Seg model wins. 
In this particular case, the L - Seg model had the advantage of being   
fully pre - trained on a much larger set of images, covering about $775 \, km^{2}$ from Romania, of higher quality and resolution (collected from Google and Bing Maps) and then fine - tuned on our Timisoara set. The LG - Seg model was only trained on Timisoara images. Qualitative results on this set are shown in \labelindexref{Figure}{img:fig_roads_results}.
The examples show the high level of difficulty posed by this challenging dataset, which we
make available for download~\footnote{https://sites.google.com/site/aerialimageunderstanding/}. 
We believe it poses a very challenging task and could help in new valuable research in 
aerial image understanding.

Our experiments on the three datasets, of different content and quality, reveal one more time the
importance of data in learning. When the structures are regular and look very similar across images, such as it is the case
with the Massachusetts Buildings, the performance reaches almost human level. However, as the variations in the data, lack of image 
quality and frequency of occlusions increase, the performance starts degrading, dropping by almost $30\%$ on the Romanian roads.
These results prove that 
aerial image understanding is far from being solved even in the context of state-of-the-art
deep networks and remains a very challenging problem.

\begin{table}
\begin{center}
{\caption{\textbf{Quantitative F - measure results on Romanian Roads Dataset}}\label{tab:Romanian_roads}}
\begin{tabular}{lcc}
\hline
\rule{0pt}{10pt}
 Method&{L-Seg}&\multicolumn{1}{c}{LG-Seg}
\\
\hline
\\[-6pt]
\quad F-measure & 66.1\% & \textbf{66.5}\% \\
\hline
\end{tabular}
\end{center}
\end{table}

\section{Detection and Counting of Romanian Houses}
\label{sec:counting_houses} 

An obvious application of building detection that is also useful in applications such as real-estate and cadaster mapping, urban planning and landscape monitoring, is the detection and counting of houses within a given area. 

For this experiment we have collected images from different areas around the city of Satu Mare, Romania, thus creating two new datasets, different from those presented in the previous sections. Besides the fact that the images were collected from rural regions with lower house densities, they were also retrieved at different spatial resolutions. For these images we have not used pixel - wise ground truth labels for training and evaluation, since these regions were not properly labeled on OSM. We refer to these datasets as \textbf{Satu Mare 1} and \textbf{Satu Mare 2}. The datasets as well as the experiments are presented next.

\subsection{Satu Mare Dataset 1}
\label{sub-sec:sm_1}

This dataset represents an aerial map of size 20000 x 20000 and spatial resolution of 0.5 x 0.5 square meters per pixel. It was divided in 400 tiles of size 1000 x 1000 pixels. The tiles were then resized with a rescale factor of 1 / 2 in order to bring the images at a resolution of $1 \, m^{2} / pixel$, closer to the one that our LG - Seg model was trained on (the European Buildings dataset). Note that the Satu Mare images were only used at test time, without any fine - tuning of the LG - Seg model. We expect that such refinement would have increased performance. However, even for this case, our results, presented next, are very promising. Also note that the houses from this region are sparsely placed, with relatively few residential areas and large vegetation regions. Also, the images are of poorer quality (see \labelindexref{Figure}{img:fig_sm_1}) than those from the European dataset that was used for training. This makes building detection a difficult task even for humans.

\fig[scale=0.17]{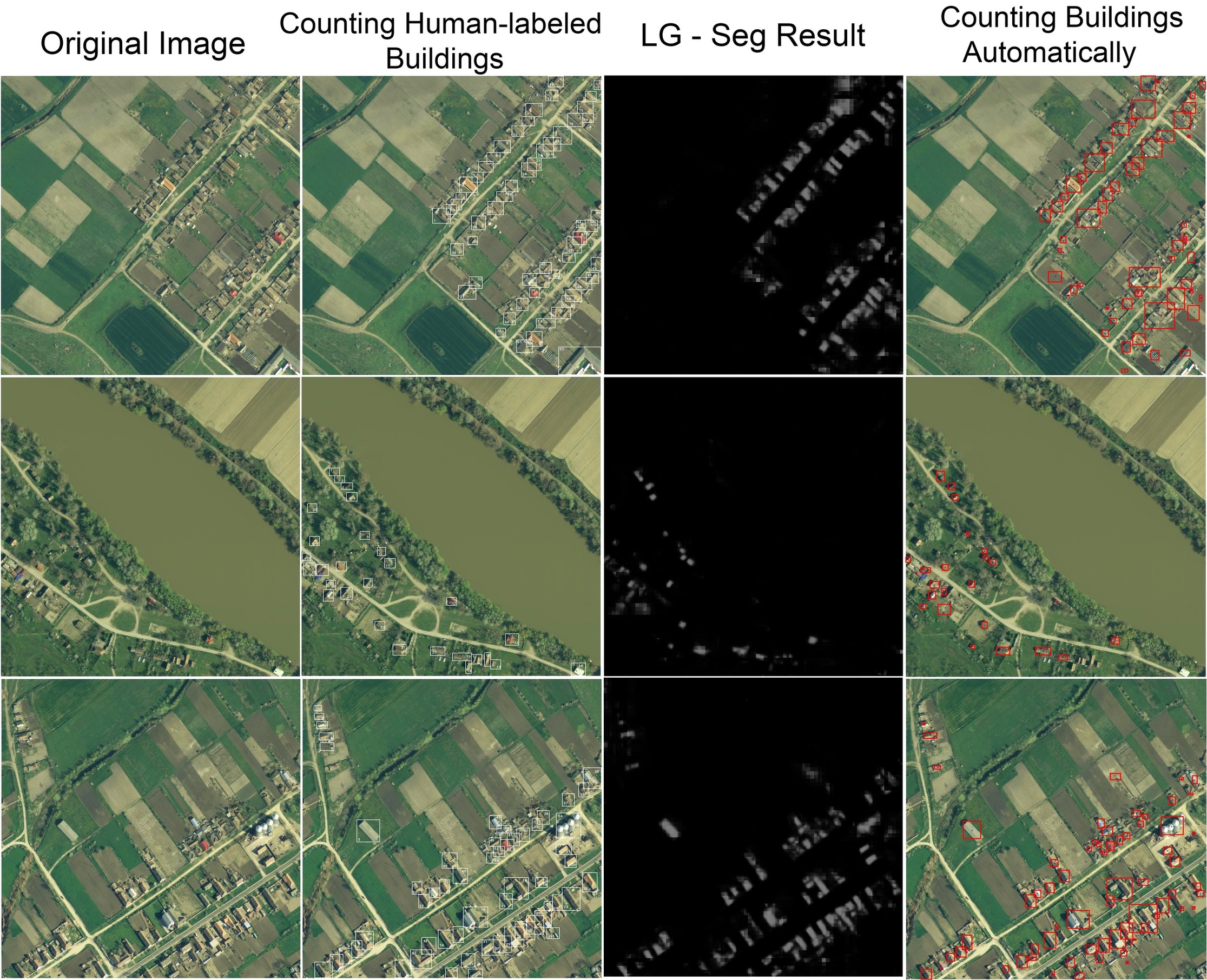}{img:fig_sm_1}{Qualitative results of building detection and house number estimation for Satu Mare Dataset 1 (spatial resolution of 0.5 $m^{2} / pixel$).} 

\subsection{Satu Mare Dataset 2}
\label{sub-sec:sm_2}

A different aerial map of size 20000 x 20000 and spatial resolution of 0.05 x 0.05 square meters per pixel was divided in 4 tiles of size 10000 x 10000. In this case we applied a rescale factor of 1 / 20 in order to bring the images to $1 \, m^{2} \ pixel$. Also different from Satu Mare 1 dataset, the buildings in Satu Mare 2 are more tightly clustered together, with a larger variation in house density. 

\fig[scale=0.17]{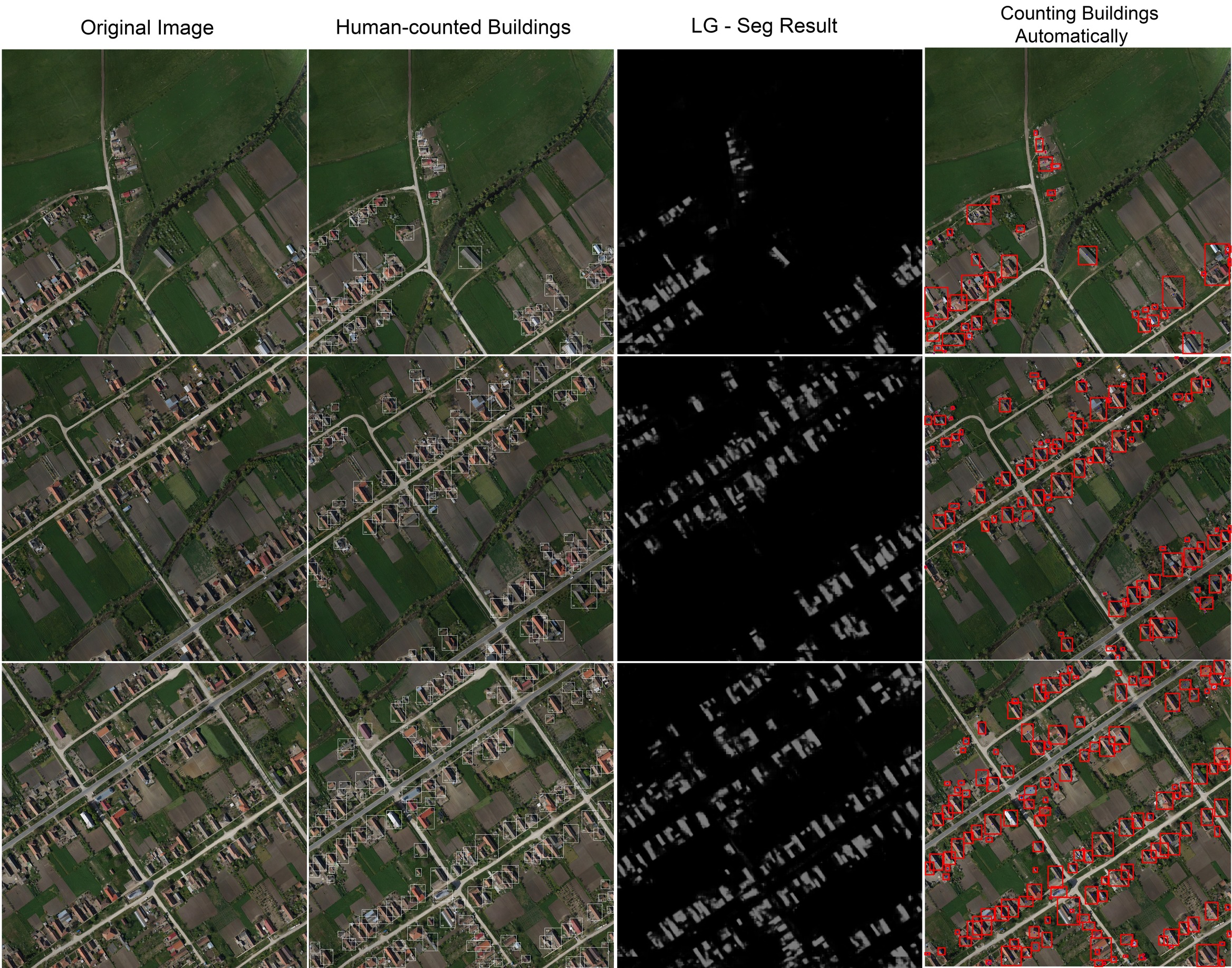}{img:fig_sm_2}{Qualitative results of building detection and house number estimation for Satu Mare Dataset 2 (spatial resolution of 0.05 $m^{2} / pixel$).} 

\subsection{Estimating the number of houses}
\label{sub-sec:no_houses}

We applied a post - processing method to the output of our model. We binarized the prediction maps by applying an optimal threshold. The value of this threshold is chosen based on our results on the European Buildings Dataset. For each image in our testing set we determined the threshold that provides the highest F - measure for that image. We then use the average over all these values, as the starting value and then varied this value in upper and lower ranges, with small steps to determine the threshold that provides the best visual results over all the testing scenarios. Continuous regions of white pixels are the buildings area that we obtain after applying the selected threshold. We apply a morphological erosion operation in order to separate closely placed buildings. Each connected component from the binary image represents a building. Counting these blobs will determine the number of buildings present in the evaluated area. Each detected building has its corresponding bounding box. Qualitative results of our method can be viewed in \labelindexref{Figure}{img:fig_sm_1} and \labelindexref{Figure}{img:fig_sm_2}.

Results are shown on two different datasets. The differences between our datasets can be easily spotted. In the first dataset we can see that the images present low building density at low resolution, whilst in the second dataset we can observe high quality images with high - density housing. For each of Satu Mare datasets we show four example results. On the first column the input RGB images, on the second column the manually - labeled bounding boxes of the buildings, on the third column the prediction map of our LG - Seg model and on the last column the result after applying our buildings counting method. Each automatically detected building has its own bounding box, in order to have a visual estimation of the surface covered by the building. The quantitative results of our building count estimation are presented in \labelindexref{Table}{tab:SM_Statistics}.

\begin{table*}[hbtp]
\centering
\footnotesize
{\caption{\textbf{House counting statistics on Satu Mare datasets.}}\label{tab:SM_Statistics}}
\begin{tabulary}{0.9\textwidth}{L|C|C|C|C|C|C|C|C|C}
\hline
\rule{0pt}{10pt}
\textbf{$SM^{1}$}&{Human Count}&{Detected Count}&{Prec}&{Rec}&{F1-Score}&{TP}&{FP}&{FN}&{Resid.}
\\
\hline
\\[-6pt]
\quad   & 156 & 137 (88\%)& 88\% & 80\% & 88.8\% & 106 & 18 & 33 & 13  \\
\hline
\textbf{$SM^{2}$}&{Human Count}&{Detected Count}&{Prec}&{Rec}&{F1-Score}&{TP}&{FP}&{FN}&{Resid.}
\\
\hline
\\[-6pt]
\quad   & 295 & 290 (98.3\%)& 90\% & 94\% & 92\% & 239 & 31 & 18 & 20  \\
\hline
\end{tabulary}
\end{table*}

Since ground truth labels were unavailable for the Satu Mare datasets, in order to evaluate the performance of our buildings counting method, we selected three examples from both of the datasets and then we manually labeled and counted the houses in these examples. 
We counted the buildings from all the three images and summed the results. The quantitative results after counting the houses for the examples in \labelindexref{Figure}{img:fig_sm_1} and \labelindexref{Figure}{img:fig_sm_2} are represented in \labelindexref{Table}{tab:SM_Statistics}. In the table we added the number of human counted houses, the true positive responses of the detector, as well as the false positives and false negatives. In some cases, a clear separation of the blobs could not be obtained. Although the bounding box was placed correctly, in an area that indeed contained the object of interest, it did not contain only an individual house, but more such objects very close to one another. Therefore we decided not to penalize the detectors response, but instead to also count these correct residential regions, that contain at least two houses. This number is marked as "Residential" in the \labelindexref{Table}{tab:SM_Statistics}.

Our house counting method provides encouraging results. For the three examples from Satu Mare Dataset 1, out of 156 human - labeled houses, our system was able to detect 106 of individual houses and 13 bounding boxes containing groups of houses. Knowing that for each residential hit there are at least two houses, we can compute an approximate value of the precision of our counting system (88\%), and recall (80\%). On the second dataset, the one with high - density housing, out of a total of 295 manually - labeled houses, there were 239 correctly detected houses, resulting in a precision of 90\% and recall of 94\%. As expected, higher image quality improves the building detection rate using our trained model. But even in low resolution conditions the system offers promising results. 

\section{Semantic segmentation of other classes}
\label{sec:other_classes}

As stated by Saito et. al ~cite{saito2015buildings} in his work, decision fusion systems have achieved accurate extraction  of terrestrial  objects  from  aerial imagery.  However, local  image  features  were  specially  designed for extracting a specific object, and the fusion techniques of multiple classifier decisions have  also  been  intended  to  be  utilized  to  extract  a  specific  object.  There are not  as  many  methods  for extracting multiple objects at the same time as for extracting each object separately, though there are many kinds of terrestrial objects in aerial imagery, and the applications cannot be achieved by extracting only one kind of object, and terrestrial objects may be correlated with each other especially in the case of buildings and roads in urban scenes. 

We have demonstrated the effectiveness of our combined network on two difficult classes, mainly on the problem of countable objects, such as buildings with different structural variations and  on  the  problem  of  road  segmentation,  considered  a  more  difficult  problem  since  there  are  fewer discriminative features to detect in the case of a continuous object. But these are not the only classes than  we  can  detect  from  aerial  images.  Besides  houses  and  roads,  we  have  encountered  areas  of  meadows, forest or water in aerial imagery. Therefore we have extended our recognition task to other classes.

\subsection{Meadows segmentation}
\label{sub-sec:meadows}

Meadow segmentation has multiple use cases in rural areas. For example, crop field statistics can be a valuable tool for farmers - adjusting the fertilizer amount, predicting the yield, determining which soil type is suitable for a specific crop. Our meadow class is general - it encompasses all types of farmland and grassland. The promising results achieved on our data opens future research possibilities for labeling specific crop types and design a yield maximization solution.

In the case of meadows segmentation from aerial images, we have extracted data in the same manner as previously stated in the case of  buildings and roads using Bing and Google RGB satellite images  aligned with ground truth labels from Open Street Map. We trained the locally informed VGG - Net with  approximately 1.5 million patches extracted from 157 training images of size 1550 x 1600. For the validation of our model we used 15 high - resolution images. As a quantitative evaluation of our model we computed the mean F - measure score and the precision – recall curve over our testing set consisting of 212 images using a set of thresholds from 0.05 to 0.95 with a step of 0.05 and obtained great results on our data (\labelindexref{Figure}{img:fig_meadows}). We obtain a mean F - measure value of 0.98648 with the relaxed version of the metric. 
 
\fig[scale=0.36]{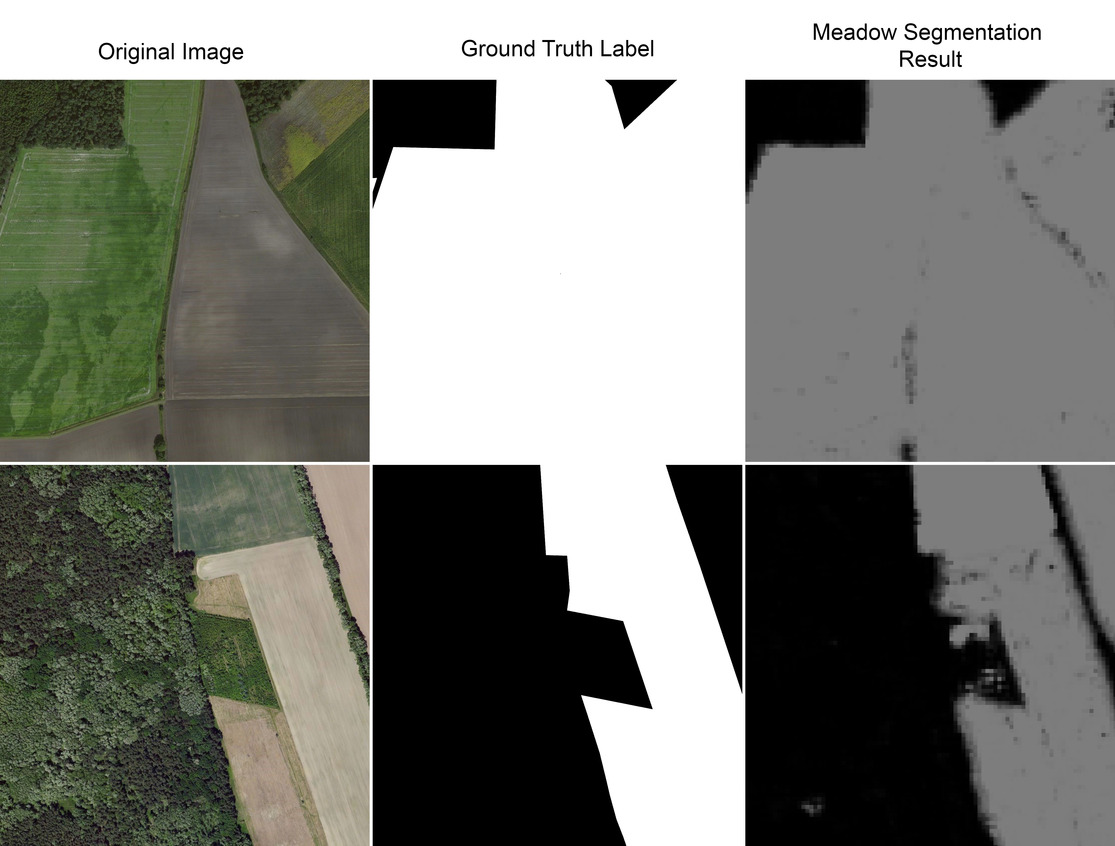}{img:fig_meadows}{ Results of our L - Seg model trained on patches with meadows as the positive class.}

\subsection{Water segmentation}
\label{sub-sec:water}

We have attempted to train L - Seg for water detection. Unfortunately, the local information patch was not sufficient for a successful train of the L - Seg. Even after significant changes for the input dataset, we have noticed that there was insufficient information in the local patch to make the model converge.

We have found a workaround for this problem - we used a larger patch of information and trained G - Seg instead. This particular model was  trained  on  approximately  1.1 million  patches  extracted  from  170  training  images  and  26  validation images.  We have used  the  same  evaluation  metric,  mean  F - measure over  the  whole  testing  set of  317 images. Results are presented in \labelindexref{Figure}{img:fig_water}. Mean F - measure has a value of 0.94377.

\fig[scale=0.43]{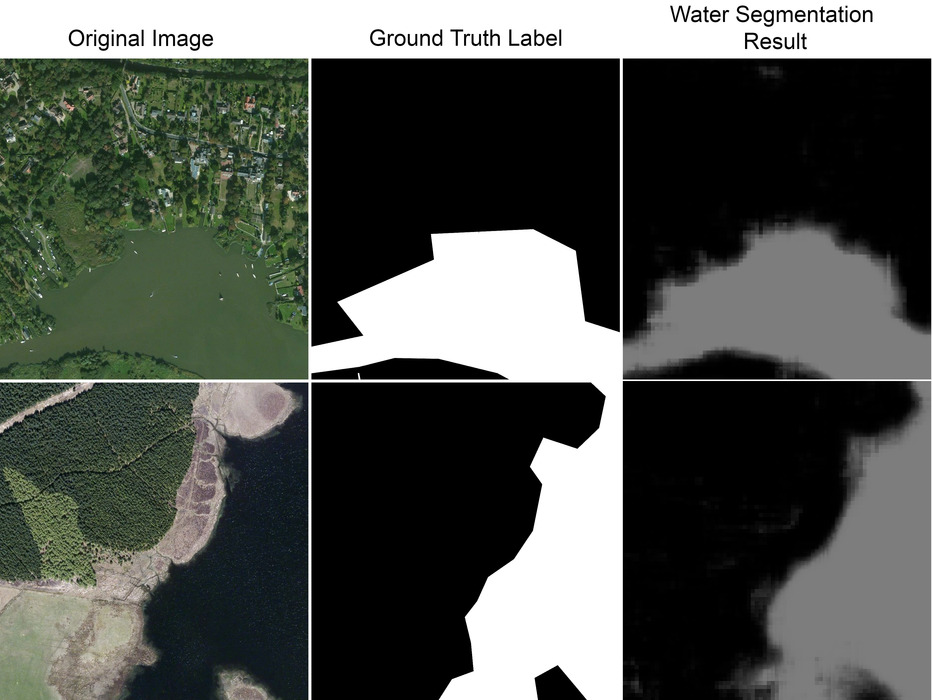}{img:fig_water}{ Results of our model on the water segmentation problem.} 

Water was trained as a general class (including rivers, lakes and sea). Given the promising results, we also attempt to classify specific water areas, in an attempt to differentiate white water from lakes or flood water. This could be a valuable tool for disaster response.

\subsection{Forest segmentation}
\label{sub-sec:forestry}

We also trained L - Seg on the task of forest segmentation. In this case, the training set was composed out of 200 images, validation set out of 30 images and the testing set out of 316 images. 

As can be easily noticed from \labelindexref{Figure}{img:fig_forest}, the network successfully manages to label dense forest, as well as isolated tree clusters. Again, the forest imagery used consisted of a mix of tree types (deciduous, conifers), all of them pictured in summer. The value of the mean F - measure in this case is 0.949239.

\fig[scale=0.36]{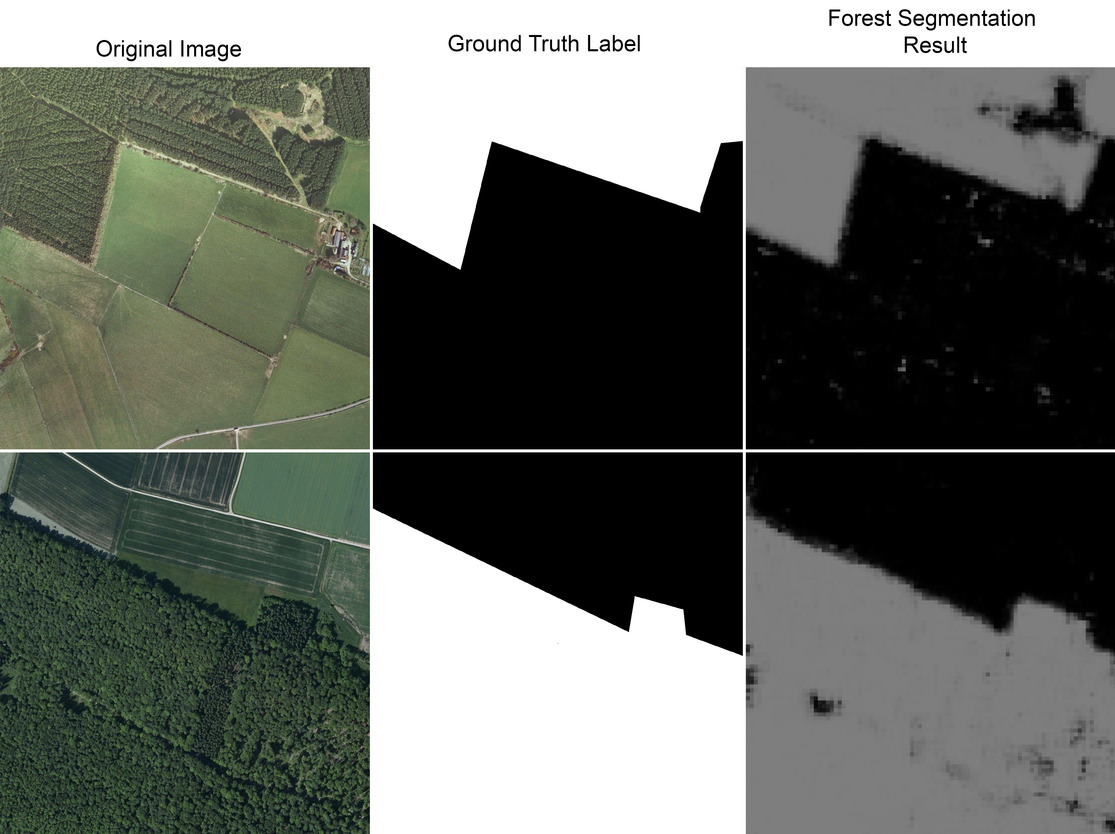}{img:fig_forest}{ Results of on semantic segmentation of forest regions.} 

We believe we could further enhance our network for the detection of specific tree species and season dependent foliage type (e.g., winter ash tree cluster versus summer ash tree cluster). We have also noticed that a tree density measure can be easily derived from our results. Therefore, our forestry information could be used for logging, landscaping or deforestation detection.

\section{Local - global complementarity}
\label{sec:local_global_complementarity}

\fig[scale=0.77]{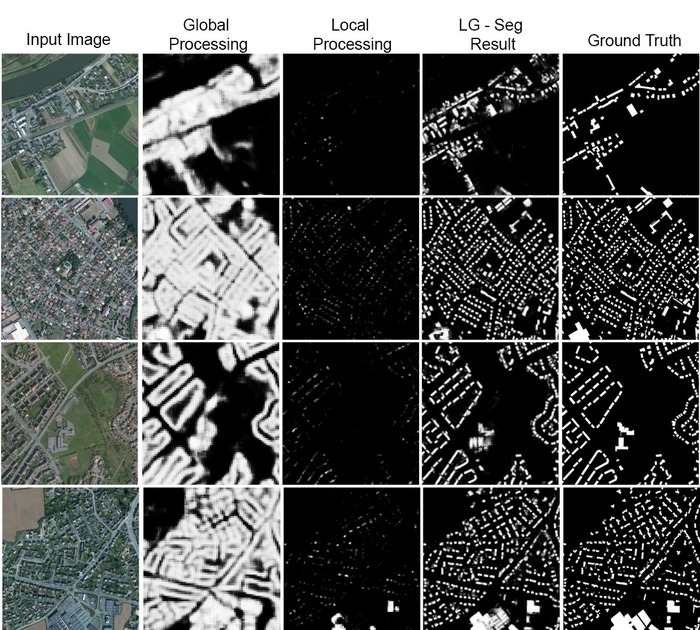}{img:fig_complementarity}{Local - global complementarity learned by the two pathways of our trained LG - Seg model.} 

One question that arises in our experiments with the dual - stream architecture is what are the two pathways learning?  What is their individual role in the combined output? Our intuition was that they learn complementary ways of processing data.
We intentionally chose two different types of networks with different image region sizes as input,
in order to encourage different learning
along the two pathways. We opted for two subnets with complementary ways of "seeing" the scene -  similar to the initial CNNs, discussed in \labelindexref{Section}{sec:initial-experiments}, one for
residential area classification (RA - Seg) and the other for local detection and segmentation of buildings (L - Seg).

We designed a set of experiments in order to better understand the role of each subnet. After training the full LG - Net, we performed the following: first, we ran the model over the test images by providing the local pathway with the correct image input, but giving a blank image to the global pathway. The blank image was the average of the original input image, each RGB channel was averaged separately. 
Then, we performed the opposite experiment and switched the inputs, by giving the original image to the global subnet and blank images to the local one. The idea was to see how, in the fully trained model, each path contributes to the final decision. 

The results, presented in \labelindexref{Figure}{img:fig_complementarity} are both very interesting and satisfying. When provided with information for local processing only, the network responds only to small buildings with very clear structure, having crisp, very local responses over individual houses or buildings. 
On the other hand, when given information only to the global subnet, the network produced a result that was closer to a soft residential area segmentation, in which individual buildings were undistinguishable from each other (a result, very close, but of higher quality than our initial residential area detection based on the same adjusted AlexNet architecture). 

In the experiments presented in \labelindexref{Figure}{img:fig_complementarity}, we aim to find what the two pathways have learned. The second column shows results when only the global pathway is fed with real image signal, the other being given blank image as input. The third column shows the opposite case, when only the local pathway is given real information. The fourth column presents the output of the network running normally, with both pathways having image input. Note that the global subnet learns to detect residential areas similar to our initial classifier for such regions. We also tested our complementarity experiment on the same image as in \labelindexref{Figure}{img:fig_intuition} \textbf{B}. Note that the residential area segmentation produced by the LG - Seg is superior to the one produced by the RA - Seg classifier, even though in the case of LG - Seg it was not asked to learn about residential areas. Also note that the local pathway focuses only on small, detailed structures. The imbalance between the energy levels of the outputs is due to the fact that one of the inputs is blank, thus unbalancing the way energy flows at the highest fully - connected layers. The results also suggest that the two pathways have roles of both reinforcement and inhibition. For example the local pathway will inhibit the global positive outputs in spaces  between buildings, whereas the global pathway will inhibit the local hallucinations in areas of low residential density. We can safely conclude that the two pathways work in complementarity.

While not focusing on fine - tuning the exact filter sizes and number of layers, we believe that our experiments have shown that our local - global strategy, by considering differently the two pathways (with architectures of different depths and filter sizes) is able to outperform each stream in isolation and learn local - global complementarity by itself. What makes these results really interesting is the fact that we did not tell these two pathways to take these different roles - all we did is choose two different architectures, gave them two different image sizes as input and let them learn, from random initializations, by themselves within the joint network. Complementarity, which was our main goal when starting this work, was learned automatically by our model from scratch.

\chapter{Conclusions and future work}
\label{chapter:conclusions}

We  have  studied  different  ways  of  combining  local  appearance  and  global  contextual information for semantic segmentation in aerial images. After testing initial simpler models that proved the  usefulness  of  reasoning  about visual  context  in  object  detection  and  segmentation,  we  have proposed a novel dual local - global network which learns completely by itself to look at objects from two complementary perspectives. One can argue that the performance of our network can be exceeded with the use of a deeper neural network,  providing  even  more  abstract  features,  but our local - global  approach  has  several advantages  over  a  single  deep  network.  By  functioning  simultaneously  the local  and  global  pathways  can learn  complementary  aspects  about  the  scene  and  naturally  include contextual  information.  The  dual - stream  net  learned  quicker  and  better  than  both  the  local  and  the  global  pathways  trained separately. Therefore,  we  expect  that  an  even  deeper  architecture,  would  have  taken  substantially  longer  to  train, while being  more  vulnerable  to  over - fitting. While  not  focusing  on  fine - tuning  the  exact  filter  sizes  and number of layers, we believe that the experiments that we have conducted have shown that our two pathways strategy is able to outperform each stream in isolation and learn local - global complementarity by itself. We prove the generalization capacity of our model on the roads dataset in which we trained on one city and tested on the other.

When  given  the  task  of  segmentation  of  buildings the  network  learns  to  treat  each  pixel, in parallel, both as a part of a building and as a part of a larger residential area. It also learns to combine the two reciprocal views in a harmonious way during the final layers of processing, before providing the final result. We stress out that the qualitative difference between our LG - Seg approach and the L - Seg model  is  clearly  visible  on  the  output  map  in  non - residential  areas  where  the  single  net  hallucinates houses. Numerically, the false positives do not affect the average F - measure by a large value. While the improvement on the Massachusetts Buildings Dataset between 2013 and 2015 was less than 0.5\%, we brought a significant improvement, from 92.3\% to 94.2\%.

While  context  should  also  be  studied  in  the  larger  spatio - temporal  domain,  in  the  image domain,  it  can  be  taken  into  consideration  only  by  looking  at  larger  spatial  support.  Unlike  other methods  using  dual - stream  nets,  we  treated  differently  the  local - global  pathways,  using  two different architectures - one deep and narrow, with small filters and the other shallower and with larger filters. The two pathways were encouraged to treat the two spatial supports differently, one focusing on small  structures  while  the  other  on  residential  areas. We  believe  that  this  is  the  main  scientific contribution of our paper, which brings new and valuable insights into at least one important aspect of context in visual recognition, that of local - global complementarity.

After  reading  various  contributions  and  existing  methods  concerning  the  problem  of  object detection in general and building detection in particular, some state-of-the-art approaches were chosen as  reference  based  on  the  results  they  obtain  using  various  techniques, from  basic  appearance methods, statistical models, hierarchical and contextual representation, to machine learning algorithms, and even a deeper understanding using deep neural networks. In this particular case, the object detection problem is considered a learning problem. First, the system learns various object properties and features extracted using descriptors from the set of training data. Second, the resulting model is applied on new data in order to detect the object class and localize the object in the scene. 

\fig[scale=0.195]{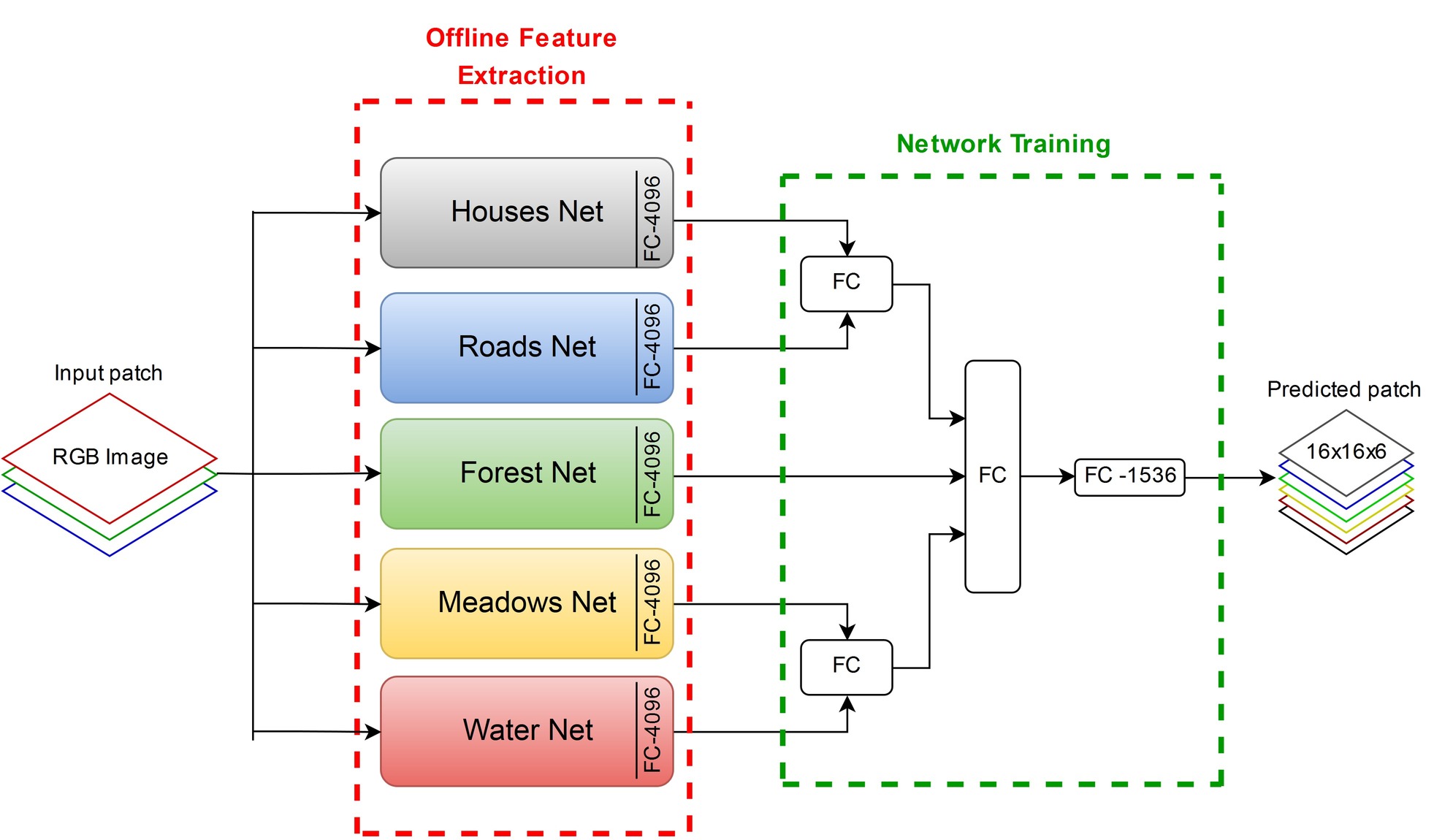}{img:fig_future_work}{ Our approach to combine all the features from different “expert” branches of processing to a higher level of feature processing in order to compute a multi - channel, multi - class segmentation of objects in aerial images.} 

We  plan  to  focus  our  future  efforts  in  two directions: improving accuracy and speed. For accuracy improvement, we plan to further investigate the use  of  spatial context  in  our  problem, along with the importance of class correlation for other objects. In  order  to  address  this  problem,  we  plan  to  extend our model to detect  multiple  classes  commonly  seen  among  aerial  images  (such  as  roads,  forests,  water  or  cars)  at once and discover how these classes can help each other to improve the final result. 

Our experiments on the roads dataset also emphasize how difficult aerial image understanding still is,  even  for  high  performance,  state-of-the-art  deep  neural  networks,  especially  in  cases  of  poor  lighting, low image quality, occlusion and high degree of variations in objects structure and shape. We  believe  that  these  limitations  will  be  overcome  by  the  usage  of  context  at  even  higher  levels  of  abstraction and reasoning. The segmentation experiments that we have conducted so far give us a great idea how to combine all these networks to perform a higher level feature processing for an even challenging task. We present  a new way of combining multiple sources of knowledge extracted from the same image but combining several “expert networks” on individual tasks in order to have a multi - label prediction map (\labelindexref{Figure}{img:fig_future_work}). Each network will process the image in the way it knows best, each branch will be responsible for extracting relevant features of the object is was trained to detect. The features extracted from each branch  are  then combined  into  further  levels of  processing,  combining  the  features  at a  higher  level. This architecture will be trained using patches in the same manner as our other models and will be able to predict multi - label patches of size 16 x 16.

The  previously  detailed methods are  intended  to  improve  the  way  flying  robots  perceive  the overflown area. The theory and practice of mobile flying robots is an emerging field, the market share of drones  will  increase  and  also  many  applications  that  before  were  neither  available  nor  possible  will become  now  achievable. Localization  and  Navigation  by  Automatic  Image  Alignment  between  Image and  Model  where  an  image  seen  from  flight,  after  feature matching,  will  be  align  to  the map model, would  allow  various  map  statistics  and  applications  never  before  possible,  such  as  environment monitoring, on-the-fly  flood / earthquake / fire  damage  evaluation  or  even  unauthorized  buildings detection. Consequently, we see our work as having the potential to influence future  research that will shed new light on the understanding of context in vision.


\bibliography{src/main}
\bibliographystyle{plain}

\printindex

\end{document}